\documentclass[journal]{IEEEtran}
\ifCLASSINFOpdf
\else
\fi
\makeatletter

\newcommand{\Rmnum}[1]{\expandafter\@slowromancap\romannumeral #1@}
\makeatother
\usepackage{times}
\usepackage{epsfig}
\usepackage{graphicx}
\usepackage{amsmath}
\usepackage{amssymb}
\usepackage{multirow}
\usepackage{tabularx}
\usepackage{caption}
\usepackage{subcaption}
\usepackage{bm}
\usepackage{booktabs}
\usepackage{float}
\usepackage{array}
\usepackage{epstopdf}
\usepackage{slashbox}
\usepackage[dvipsnames]{xcolor}
\usepackage{hyperref}
\usepackage{cleveref}
\usepackage{indentfirst}
\newcommand{\tabincell}[2]{\begin{tabular}{@{}#1@{}}#2\end{tabular}}

\ifCLASSOPTIONcompsoc
  \usepackage[nocompress]{cite}
\else
  \usepackage{cite}
\fi
\newcommand\myshade{85}
\colorlet{mylinkcolor}{violet}
\colorlet{mycitecolor}{purple}
\colorlet{myurlcolor}{Aquamarine}

\hypersetup{
  linkcolor  = mylinkcolor!\myshade!black,
  citecolor  = mycitecolor!\myshade!black,
  urlcolor   = myurlcolor!\myshade!black,
  colorlinks = true,
}

\begin{document}
%
\title{Manifold Criterion Guided Transfer Learning via Intermediate Domain Generation}
%
%
%
%

\author{Lei Zhang,~\IEEEmembership{Senior Member,~IEEE,~}
        Shanshan Wang,
        Guang-Bin Huang,~\IEEEmembership{Senior Member,~IEEE,~}
        Wangmeng Zuo,~\IEEEmembership{Senior Member,~IEEE,~}
          Jian Yang,~\IEEEmembership{Member,~IEEE,~}
           David Zhang,~\IEEEmembership{Fellow,~IEEE}
\thanks{This work was supported by the National Science Fund of China under Grants (61771079, 91420201 and 61472187), Chongqing Science and Technology Project (No. cstc2017zdcy-zdzxX0002, cstc2018jcyjAX0250), the 973 Program No.2014CB349303, and Program for Changjiang Scholars. \textit{(Corresponding author: Lei Zhang)}
}
\IEEEcompsocitemizethanks{\IEEEcompsocthanksitem L. Zhang and S. Wang are with the School of Microelectronics and Communication Engineering, Chongqing University,  Chongqing 400044,China.
\ (E-mail: leizhang@cqu.edu.cn, wangshanshan@cqu.edu.cn).
\IEEEcompsocthanksitem G.B. Huang is with School of Electrical and Electronic Engineering, Nanyang Technological University, Singapore.
 \ (E-mail: egbhuang@ntu.edu.sg).
\IEEEcompsocthanksitem W.M. Zuo is with School of Computer Science and Technology, Harbin Institute of Technology, Harbin, China.
 \ (E-mail: wmzuo@hit.edu.cn).
\IEEEcompsocthanksitem J. Yang is with School of Computer Science and Technology, Nanjing University of Science and Technology, China.
 \ (E-mail: csjyang@njust.edu.cn).
\IEEEcompsocthanksitem D. Zhang is with School of Science and Engineering, Chinese University of Hong Kong (Shenzhen), Shenzhen, China.
 \ (E-mail: davidzhang@cuhk.edu.cn).}
}

%
%

\markboth{IEEE TRANSACTIONS ON NEURAL NETWORKS AND LEARNING SYSTEMS,~Vol.~XX, No.~XX, May~2018}%
{Shell \MakeLowercase{\textit{et al.}}: Bare Demo of IEEEtran.cls for Computer Society Journals}
%



\maketitle

\begin{abstract}

In many practical transfer learning scenarios, the feature distribution is different across the source and target domains (i.e. non-$i.i.d.$). Maximum mean discrepancy (MMD), as a domain discrepancy metric, has achieved promising performance in unsupervised domain adaptation (DA). We argue that MMD-based DA methods ignore the data locality structure, which, to some extent, would cause the negative transfer effect. The locality plays an important role in minimizing the nonlinear local domain discrepancy underlying the marginal distributions. For better exploiting the domain locality, a novel local generative discrepancy metric (LGDM) based intermediate domain generation learning called Manifold Criterion guided Transfer Learning (MCTL) is proposed in this paper. The merits of the proposed MCTL are four-fold: 1) the concept of manifold criterion (MC) is first proposed as a measure validating the distribution matching across domains, and domain adaptation is achieved if the MC is satisfied; 2) the proposed MC can well guide the generation of the intermediate domain sharing similar distribution with the target domain, by minimizing the local domain discrepancy; 3) a global generative discrepancy metric (GGDM) is presented, such that both the global and local discrepancy can be effectively and positively reduced; 4) a simplified version of MCTL called MCTL-S is presented under a perfect domain generation assumption for more generic learning scenario. Experiments on a number of benchmark visual transfer tasks demonstrate the superiority of the proposed manifold criterion guided generative transfer method, by comparing with other state-of-the-art methods. The source code is available in https://github.com/wangshanshanCQU/MCTL.
\end{abstract}




%
\IEEEpeerreviewmaketitle

\section{Introduction}

\IEEEPARstart{S}{tatistical} machine learning models rely heavily on the assumption that the data used for training and test are drawn from the same or similar distribution, i.e. independent identical distribution ($i.i.d.$). However, in real world, it is impossible to guarantee that assumption. Hence, in visual recognition tasks, classifier or model usually does not work well because of data bias  between the distributions of the training and test data\cite{Nguyen2015DASH},\cite{csurka2017domain}\cite{hoffman2012discovering},\cite{pan2010survey},\cite{Duan2012Domain},\cite{li2017domain},\cite{gopalan2014unsupervised}.
The domain discrepancy constitutes a major obstacle in training the predictive models across domains. For example, an object recognition model trained on labeled images may not generalize well on the testing images under various variations in the pose, occlusion, or illumination.
In Machine Learning this problem is labeled as domain mismatch. Failing to model such a distribution shift may cause significant performance degradation.
Also, the models trained with only a limited number of labeled patterns are usually not robust for pattern recognition tasks. Furthermore, manual labeling of sufficient training data for diverse application domains may be prohibitive. However, by leveraging the labeled data drawn from another sufficiently labeled source domain that describes related contents with target domain, establishing an effective model is possible. Therefore, the challenging objective is how to achieve knowledge transfer across domains such that the distribution mismatch is reduced.
Underlying techniques for addressing this challenge, such as domain adaptation\cite{Duan2010Visual}\cite{kulis2011you}, which aims to learn domain-invariant models across source and target domain, has been investigated.

Domain adaptation (DA)\cite{saenko2010adapting},\cite{pan2011domain},\cite{Duan2009Domain} as one kind of transfer learning (TL) perspective, addresses the problem that data is from two related but different domains\cite{baktashmotlagh2013unsupervised},\cite{ben2007analysis}. Domain adaptation establishes knowledge transfer from the labeled source domain to the unlabeled target domain by exploring domain-invariant structures that bridge different domains with substantial distribution discrepancy.
In terms of the accessibility of target data labels in transfer learning, domain adaptation methods can be divided into three categories: supervised\cite{duan2012learning},\cite{pan2008transfer}, semi-supervised\cite{li2014learning},\cite{yao2015semi},\cite{Duan2012Domain} and unsupervised\cite{long2014transfer},\cite{baktashmotlagh2014domain},\cite{long2016unsupervised}.

In this paper, we focus on unsupervised transfer learning where the target data labels are unavailable in transfer model learning phase. Unsupervised setting is more challenging due to the common data scarcity problem. In unsupervised transfer learning\cite{ganin2014unsupervised}, Maximum Mean Discrepancy (MMD)\cite{gretton2007kernel} is widely used and has achieved promising performance. MMD, that aims at minimizing the domain distribution discrepancy, is generally exploited to reduce the difference of conditional distributions and marginal distributions across domains by utilizing the unlabeled domain data in a Reproducing Kernel Hilbert Space (RKHS). Also, in the framework of deep transfer learning\cite{hu2015deep}, MMD-based adaptation layers are further integrated in deep neural networks to improve the transferable capability between the source and domains\cite{long2015learning}.

\begin{figure}[t]
\centering
  \includegraphics[width=1\linewidth]{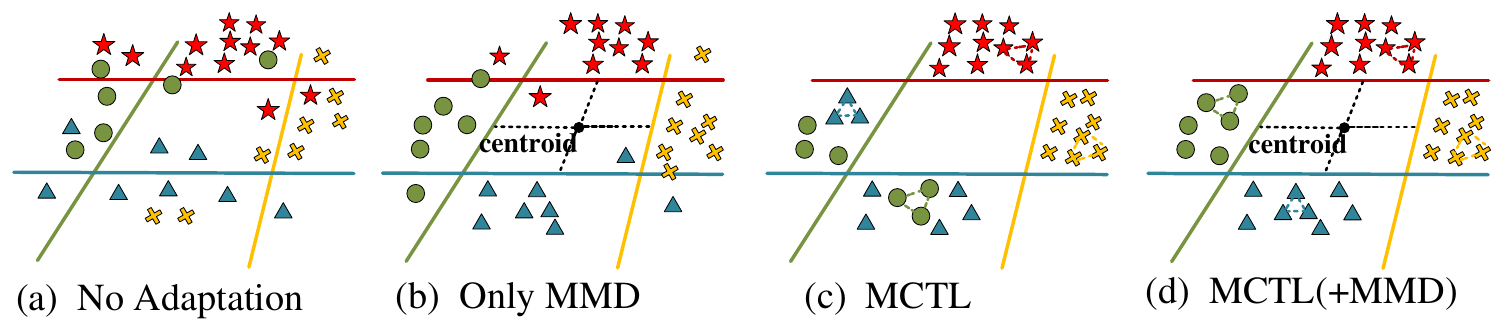}
   \caption{ Motivation of MCTL. The lines represent the classification boundary of source domain. The centroid represents the geometric center of all data points.}
   \label{fig1}
\end{figure}

MMD actually acts as a discrepancy metric or criterion to evaluate the distribution mismatch across domains and works well in aligning the global distribution. However, it only considers the domain discrepancy and generally ignores the intrinsic data structure of target domain, e.g. local structure just as Fig.\ref{fig1}(b). It is known that geometric structure is indispensable for domain distance minimization, which, thus, can well exploit the internal local structure of target data. Particularly, in unsupervised learning, the local structure of target data often plays a more important role than the global structure. This is originated from manifold assumption that the data with local similarity is with similar labels. Motivated by manifold assumption, a novel manifold criterion (MC) is proposed in our work, which is similar but very different from conventional manifold algorithms that the MC actually acts as a generative transfer criterion for unsupervised domain adaptation.

Intuitively, we hold the assumption that if a new target domain can be automatically generated by using the source domain data, the domain transfer issue can be naturally addressed. To this end, a criterion that measures the generative effect can be explored. In this paper, considering the locality property of target data, we wish that the generative target data should hold similar local structure with the true target domain data. Naturally, motivated by manifold assumption\cite{hoffman2014continuous}, an objective generative transfer metric, manifold criterion (MC), is proposed. Suppose that two samples $x_{i}$ and $x_{j}$ in target domain are close to each other, and if the generative target sample $x_{i}^g$ by using the source data is also close to $x_{j}$, we recognize that the generated intermediate domain data shares similar distribution with the target domain. This is the basic idea of our generative transfer learning in this paper.

But how to construct the generative target domain? From the perspective of manifold learning, we expect that the new target data is generated by using a locality structure preservation metric. This idea can be interpreted under the commonly investigated case of independent identically distribution ($i.i.d.$) that the affinity structure in high-dimensional space can still be preserved in some projected low-dimensional subspace (i.e. manifold structure embedding). In general, the internal intrinsic structure can remain unchanged by using graph Laplacian regularization\cite{Yang2017EPR}, which reflects the affinity of the raw data.

Specifically, with the proposed manifold criterion, a $\bf{M}$anifold $\bf{C}$riterion guided $\bf{T}$ransfer $\bf{L}$earning (MCTL) is  proposed, which aims to pursue a latent common subspace via a projection matrix $\bm{\mathcal {P}}$ for source and target domain. In the common subspace, a generative transfer matrix $\bm{\mathcal {Z}}$ is solved by leveraging the source domain data and the MC generative metric, for a new generative data that holds similar marginal distribution with source data in a unsupervised manner. The findings and analysis show that the proposed manifold criterion can be used to reduce the local domain discrepancy.

Additionally, in MCTL model, the embedding of low-rank constraint (LRC) on the transfer matrix ensures that the data from source domains can be well interpreted during generation, which can show an approximated block-diagonal property. With the LRC exploited, the local structure based MC can be guaranteed as we wish without distortion\cite{kanamori2009efficient}.

The idea of our MCTL is described in Fig.\ref{fig2}. In summary, the main contribution and novelty of this work are fourfold:
\begin{itemize}
\item We propose a unsupervised manifold criterion generative transfer learning (MCTL) method, which aims to generate a new intermediate target domain that holds similar distribution with true target data by leveraging source data as basis. The proposed manifold criterion (MC) is modeled by a novel local generative discrepancy metric (LGDM) for local cross-domain discrepancy measure, such that the local transfer can be effectively aligned.

\item In order to keep the global distribution consistency, a global generative discrepancy metric (GGDM), that offers a “linear” method to compare the high-order statistics of two distributions, is proposed to minimize the discrepancy between the generative target data and the true target data. Therefore, the local and global affinity structures across domains are simultaneously guaranteed.

\item For improving the correlation between the source data and the generative target data, LRC regularization on the transfer matrix $\bm{\mathcal {Z}}$ is integrated in MCTL, such that the block-diagonal property can be utilized for preventing the domain transfer from distortion and negative transfer.

\item Under the MCTL framework, for a more generic case, a simplified version of MCTL (i.e. MCTL-S) method is proposed, which constrains that the generative data should be seriously consistent with the target domain in a simple yet generic manner. Interestingly, with this constraint, the LGDM loss in MCTL-S is naturally degenerated into a generic manifold regularization.
\end{itemize}

The remainder of this paper is organized as follows. In Section II, we review the related work in transfer learning. In Section III, we present the preliminary idea of the proposed manifold criterion. In Section IV, the proposed MCTL method and optimization are formulated. In Section V, the simplified version of MCTL is introduced and preliminarily analyzed. In Section VI, the classification method is described. In Section VII, the experiments in cross-domain visual recognition are presented. The discussion is presented in Section VIII. Finally, the paper is concluded in Section IX.

\section{Related Work}
\subsection{Shallow Transfer Learning}

A lot of transfer learning methods are proposed to tackle heterogeneous domain adaptation problems. Generally, these methods can be divided into three categories in the follows. 

\textbf{Classifier based approaches}. A generic way is to directly learn a common classifier on auxiliary domain data by leveraging a few labeled target data. Yang et al.\cite{yang2007cross} proposed an adaptive SVM (A-SVM) to learn a new target classifier $f^T(x)$ by supposing that $f^T(x)= f^S(x) + \Delta f(x)$, where the classifier $f^S(x)$ is trained with the labeled source samples and $\Delta f(x)$ is the perturbation function. Bruzzone et al.\cite{Bruzzone2010Domain} developed an approach to iteratively learn the SVM classifier by labeling the unlabeled target samples and simultaneously removing some labeled samples in the source domain. Duan et al.\cite{Duan2010Visual} proposed an adaptive multiple
kernel learning (AMKL) for consumer video event recognition from annotated web videos. Also, a domain transfer MKL (DTMKL)\cite{Duan2012Domain}, which learn a SVM classifier and a kernel function simultaneously for classifier adaptation. Zhang et al.\cite{eda2016zhang} proposed a robust classifier transfer method (EDA) which was modelled based on ELM and manifold regularization for visual recognition.

\textbf{Feature augmentation/transformation based approaches}. Li et al.\cite{li2014hfa} proposed a heterogeneous feature augmentation (HFA)which tends to learn a transformed feature space for domain adaptation. Kulis et al.\cite{kulis2011you} proposed an asymmetric regularized cross-domain transform (ARC-t) method for learning a transformation metric. In \cite{Hoffman2014Asymmetric}, Hoffman et al. proposed a Max-Margin Domain Transforms (MMDT) which a category specific transformation was optimized for domain transfer. Gong et al. proposed a Geodesic Flow Kernel (GFK)\cite{Gong2012Geodesic} method which integrates an infinite number of linear subspaces on the geodesic path to learn the domain-invariant feature representation. Gopalan et al.\cite{Gopalan2011Domain} proposed an unsupervised method (SGF) for low dimensional subspace transfer in which a group of subspaces along the geodesic between source and target data is sampled, and the source data is projected into the subspaces for discriminative classifier learning. An unsupervised feature transformation approach, Transfer Component Analysis (TCA)\cite{pan2011domain}, was proposed to discover common features having the same marginal distribution by using Maximum Mean Discrepancy (MMD) as non-parametric discrepancy metric. MMD\cite{gretton2007kernel},\cite{gretton2012jmlr},\cite{iyer2014maximum} is often used in transfer learning. Long et al.\cite{long2013transfer} proposed a Transfer Sparse Coding (TSC) approach to construct robust sparse representations by using empirical MMD as the distance measure. The Transfer Joint Matching (TJM) proposed by Long et al.\cite{long2014transfer} tends to learn a non-linear transformation by minimizing the MMD based distribution discrepancy.

\textbf{Feature representation based approaches}. Different from those methods above, domain adaptation is achieved by representing across domain features. Jhuo et al.\cite{jhuo2012robust} proposed a RDALR method, in which the source data is reconstructed with target domain by using low-rank modeling. Similarly, Shao et al.  \cite{shao2014generalized} proposed a LTSL method by pre-learning a subspace using PCA or LDA, then low-rank representation across domain is modeled. Zhang et al. \cite{zhang2016lsdt},\cite{Zhang2016Discriminative} proposed Latent Sparse Domain Transfer (LSDT) and Discriminative Kernel Transfer Learning (DKTL) methods for visual adaptation, by jointly learning a subspace projection and sparse reconstruction across domain. Further, Xu et al. \cite{Xu2015} proposed a DTSL method, which combines the low-rank and sparse constraint on the reconstruction matrix.

In this paper, the proposed method is different from the existing shallow transfer learning methods that a generative transfer idea is motivated, which tends to achieve domain adaptation by generating an intermediate domain that has similar distribution with the true target domain.

\subsection{Deep Transfer Learning}
Deep learning, as a data-driven transfer learning method, has witnessed a great achievements in many fields\cite{tzeng2015simultaneous},\cite{glorot2011domain},\cite{oquab2014learning},\cite{xie2015transfer}. However, when solving domain data problems by deep learning technology, massive labeled training data are required. For the small-size tasks, deep learning may not work well. Therefore, deep transfer learning methods have been studied.

Donahue et al.\cite{donahue2014decaf} proposed a deep transfer method for small-scale object recognition, and the convolutional network (AlexNet) was trained on ImageNet. Similarly, Razavian et al.\cite{sharif2014cnn} also proposed to train a network based on ImageNet for high-level feature extractor. Tzeng et al.\cite{tzeng2015simultaneous} proposed a DDC method which simultaneously achieves knowledge transfer between domains and tasks by using CNN. Long et al.\cite{long2015learning} proposed a deep adaptation network (DAN) method by imposing MMD loss on the high-level features across domains. Additionally, Long et al.\cite{long2016unsupervised} also proposed a residual transfer network (RTN) which tends to learn a residual classifier based on softmax loss. Oquab et al.\cite{oquab2014learning} proposed a CNN architecture for middle level feature transfer, which is trained on large annotated image set. Additionally, Hu et al.\cite{hu2015deep} proposed a non-CNN based deep transfer metric learning (DTML) method to learn a set of hierarchical nonlinear transformations for achieving cross-domain visual recognition.

Recently, GAN inspired adversarial domain adaptation has been preliminarily studied. Tzeng et al. proposed a novel ADDA method \cite{Tzeng2017Adversarial} for adversarial domain adaptation, in which CNN is used for adversarial discriminative feature learning, and achieves the state-of-the-art performance.

In this work, although the proposed MCTL method is a shallow transfer learning paradigm, the competitive capability comparing to these deep transfer learning methods has been validated on the pre-extracted deep features.

\subsection{Differences Between MCTL and Other Reconstruction Transfer Methodologies}

The proposed MCTL is partly related by reconstruction transfer methods, such as DTSL\cite{Xu2015}, LSDT\cite{zhang2016lsdt} and LTSL\cite{shao2014generalized}, but essentially different from them. These methods aim to learn a common subspace where a feature reconstruction matrix between domains is learned for adaptation. Sparse reconstruction and low-rank based constraints were considered, respectively. Different from reconstruction transfer, the proposed MCTL is a generative transfer learning paradigm, which is partly inspired by the idea of GAN\cite{Goodfellow2014Generative} and manifold learning. The differences and relations are as follows.

\textbf{Reconstruction Transfer}. As the name implies, a reconstruction matrix is expected for domain correspondence. In LTSL, subspace projection $\bf{W}$ is  pre-learned by off-the-shelf methods such as PCA, LDA, etc. Then projected source data $\bf{WX_S}$ is used to reconstruct the projected target data $\bf{{WX}_T}$ via low-rank constraint. The subspace may be suboptimal leading to a possible local optimum of $\bf{\mathcal{Z}}$. Further, the LSDT method was proposed for realizing domain adaptation by exploiting cross-domain sparse reconstruction in some latent subspace, simultaneously. The DTSL was proposed by posing hybrid regularization of sparsity and low-rank constraints for learning a more robust reconstruction transfer matrix. Reconstruction transfer always expresses target domain by leveraging source domain, however, this expression is not accurate due to the limited number of target domain data in calculating the reconstruction error loss, and the robustness is decreased.

\textbf{Generative Transfer}. The proposed MCTL method introduces a generative transfer learning concept, which aims to realize an intermediate domain generation by constructing a Manifold Criterion loss. The motivation is that the domain adaptation problem can be solved by generating a similar domain that shares the same distribution with the true target domain. The essential differences of our work from reconstruction lie in that: (1) Domain adaptation is recognized to be a domain generation problem, instead of a domain alignment problem. (2) The manifold criterion loss is well constructed for generation, instead of the least-square based reconstruction error loss. In addition, the GGDM based global domain discrepancy loss and LRC regularization are also integrated in MCTL for global distribution discrepancy reduction and domain correlation enhancement, simultaneously.

\textbf{Similarity and Relationship}. The reconstruction transfer and generative transfer are similar and related in three aspects. (1) Both aim at pursuing a more similar domain with the target data by leveraging the source domain data. (2) Both are unsupervised transfer learning, which do not need the data label information in domain adaptation. (3) Both have similar model formulation and solvers for obtaining the domain correspondence matrix and transformation.


\begin{figure}[t]
\centering
  \includegraphics[width=1\linewidth]{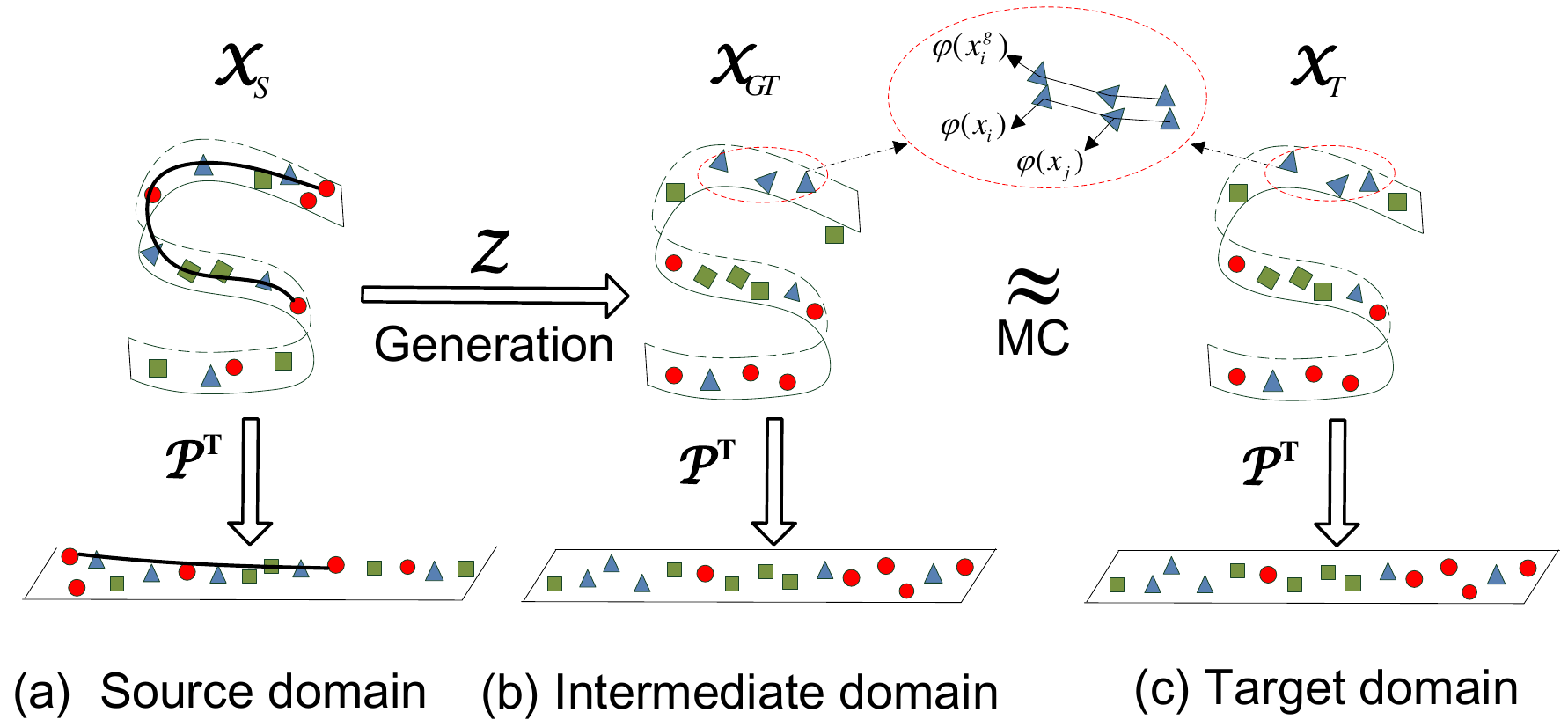}
   \caption{ Illustration of the proposed Manifold Criterion Guided Transfer Learning (MCTL). (a) represents the source domain $\bm{\mathcal{X}}_{S}$ which is used to generate an intermediate target domain $\bm{\mathcal{X}}_{GT}$ shown as (b), that is similar to the true target domain $\bm{\mathcal{X}}_{T}$ shown in (c). The intermediate domain generation is carried out by the learned generative matrix $\bm{\mathcal {Z}}$ based on the manifold criterion (MC) in an unsupervised manner. MC interprets the distribution discrepancy, which implies that if the local discrepancy is minimized, the distribution consistency is then achieved. Further, a projection matrix $\bm{\mathcal {P}}$ is learned for domain feature embedding. Notably, the $\varphi (.)$ is used as the implicit mapping function of data, which can be kernelized in implementation with inner product.}
   \label{fig2}
\end{figure}

\section{Manifold Criterion Preliminary}

Manifold learning\cite{baktashmotlagh2014domain},\cite{Yang2017EPR} as a typical unsupervised learning method has been widely used. Manifold hypothesis means that an intrinsic geometric low-dimensional structure is embedded in high-dimensional feature space and the data with affinity structure own similar labels. This demonstrates that manifold hypothesis works but under the data of independent identically distribution ($i.i.d.$). Therefore, we could have a try to build a manifold criterion to measure the $i.i.d.$ condition (i.e. domain discrepancy minimization) and guide the transfer learning across domains through an intermediate domain.

In this paper, manifold hypothesis is used in the process of generating domain as shown in Fig.\ref{fig2}. Essentially different from manifold learning and regularization, we propose a novel manifold criterion (MC) that is utilized as generative discrepancy metric. In semi-supervised learning (SSL), manifold regularization is often used but under $i.i.d.$ condition. However, transfer learning is different from SSL that domain data does not satisfy $i.i.d.$ condition. In this paper, it should be figure out that if the intermediate domain can be generated via the manifold criterion guided objective function, then the distribution of the generated intermediate domain and the true target domain is recognized to be matched.

The idea of manifold criterion is described in Fig.\ref{fig2}. We observe that a projection matrix $\bm{\mathcal {P}}$ is first learned for some common subspace projection, and then a generative transfer matrix $\bm{\mathcal {Z}}$ is learned for intrinsic structure preservation and distribution discrepancy minimization between the true target data and generative target data by source domain data. That is, if the generative data has similar affinity structure with the true target domain, i.e. manifold criterion is satisfied, we can have a conclusion that the generative data shares similar distribution with target domain. Notably, different from reconstruction based domain adaptation methods, in this work, we tend to generate an intermediate domain by leveraging source domain, i.e. generative transfer instead of reconstruction transfer.

Moreover, we show Fig.\ref{fig1} to imply that MC (local) and MMD (global) can be jointly considered in transfer learning models.
Frankly, the idea of this paper is intuitive, simple and easy to follow. The key point lies in that how to generate the intermediate domain data such that the generated data complies with manifold assumption originated from the true target domain data. If the manifold criterion is satisfied (i.e. $i.i.d.$ is achieved), then domain adaptation or distribution alignment is completed, which is the principle of MCTL.

\section{MCTL: Manifold Criterion Guided Transfer Learning}
\subsection{Notations}
In this paper, source and target domain are defined by subscript “$S$” and “$T$”. Training set of source and target domain are defined as $\varphi {\bf{(}}{{\bf{X}}_S}{\bf{)}} \in {{\bf{R}}^{m\times {n_S}}}$  and $\varphi {\bf{(}}{{\bf{X}}_T}{\bf{)}} \in {{\bf{R}}^{m\times {n_T}}}$. $\varphi {\bf{(}}{{\bf{X}}_{{\rm{G}}T}}{\bf{)}} \in {{\bf{R}}^{m\times {n_T}}}$  denotes generative target domain, where $\varphi$ denotes an implicit but generic transformation, $m$ denotes dimensionality, $n_{S}$ and $n_T$ denote the number of samples in source and target domain, respectively. Let ${\bf{X}} = [{{\bf{X}}_S}, {{\bf{X}}_T}]$, then $\varphi {\bf{(}}{{\bf{X}}}{\bf{)}} \in {{\bf{R}}^{m\times n}}$, where ${n = {n_S}+{n_T}}$. Let ${\bm{\mathcal {P}}} \in {{\bf{R}}^{m\times d}}$$(m\geq d)$  be the basis transformation that maps raw data space from ${\bf{R}}^m$ to a latent subspace ${\bf{R}}^d$. ${\bm{\mathcal{Z}}} \in {{\bf{R}}^{{n_S}\times{n_T}}}$ represents generative transfer matrix, $\mathbf{I}$ denotes identity matrix, $\parallel\bullet\parallel_{F}$ and $\parallel\bullet\parallel_{2}$ denote $l_{F}$-norm and $l_{2}$-norm, respectively. The superscript $\mathbf{T}$ denotes transpose operator and $\mathbf{Tr(\bullet)}$ denotes matrix trace operator.

In RKHS, the kernel Gram matrix $\bm{\mathcal{K}}$ is defined as $\begin{bmatrix}{\bf{K}}\end{bmatrix}_{i,j}=<\varphi(\mathbf{x}_i), \varphi(\mathbf{x}_j)>_\mathcal{H}=\varphi(\mathbf{x}_i)^H\varphi(\mathbf{x}_j)=k(\mathbf{x}_i,\mathbf{x}_j)$,
where $k$ is a kernel function. In the following sections, let ${\bf{{K}}}{\rm{ = }}\varphi {{\bf{(X)}}^{\bf{T}}}\varphi {\bf{(X)}}$, ${{\bf{K}}_S} = \varphi {{\bf{(X)}}^{\bf{T}}}\varphi ({{\bf{X}}_S})$ and ${{\bf{K}}_T} = \varphi {{\bf{(X)}}^{\bf{T}}}\varphi ({{\bf{X}}_T})$,
 and it is easy to get that ${\bf{K}} \in {{\bf{R}}^{n \times n}}$, ${{\bf{K}}_S} \in {{\bf{R}}^{n \times {n_S}}}$ and ${{\bf{K}}_T} \in {{\bf{R}}^{n \times {n_T}}}$.
\subsection{Problem Formulation}

In this section, the proposed MCTL method is presented in Fig.\ref{fig2}, in which the same distribution between the $\bf{G}$enerated intermediate $\bf{T}$arget domain ($D_{GT}$) and the true $\bf{T}$arget domain ($D_T$) under common subspace is what we expected. That is, the intermediate target domain is generated to share the approximated distribution as the true target domain by exploiting the proposed Manifold Criterion as domain discrepancy metric. Specifically, two generative discrepancy metrics (LGDM vs. GGDM) for measuring the domain discrepancy locally and globally are proposed. Overall, the model is composed of three items. The $1^{st}$ item is MC-based LGDM loss which is used to measure the local domain discrepancy with the manifold criterion by exploiting the locality of target data. The $2^{nd}$ item is the GGDM loss which is applied to minimize the global domain discrepancy of marginal distributions between the generated intermediate target domain and the true target domain. The $3^{rd}$ item is the LRC regularization (low-rank constraint) which is carried out to keep the generalization of $\bm{\mathcal {Z}}$. A detailed MCTL method is described in the follows.

\subsubsection{MC based Local Generative Discrepancy Metric}
The MC based local generative discrepancy metric (LGDM) loss is used to enhance the distribution consistency between source and target domain indirectly, by constraining the generative target data with manifold criterion. For convenience, $\varphi {\bf{(}}x_{GT}^p{\bf{)}}$ is defined as a sample in $\varphi ({{\bf{X}}_{GT}})$ and $\varphi {\bf{(}}x_T^q{\bf{)}}$ is defined as a sample in $\varphi ({{\bf{X}}_T})$. We claim that the distribution consistency between $\varphi ({{\bf{X}}_{GT}})$ and $\varphi ({{\bf{X}}_T})$ is achieved, i.e. domain transfer is done, only if two sets satisfy the following manifold criterion, which can be formulated as
\begin{equation}
\begin{aligned}
\mathop {LGDM({D_{GT}},{D_T})=}\limits_{} &\sum\limits_{p,q}^{{n_T}} {{W_{pq}}} \left\| {\varphi {\bf{(}}x_{GT}^p{\bf{)}} - \varphi {\bf{(}}x_T^q{\bf{)}}} \right\|_2^2\\
 &{\rm{ = }}\mathop {}\limits_{} Tr(\varphi ({{\bf{X}}_{GT}}){\bf{D}}(\varphi {({{\bf{X}}_{GT}})^{\bf{T}}}) \\
 &\hspace*{0.3cm}+ Tr(\varphi ({{\bf{X}}_T}){\bf{D}}(\varphi {({{\bf{X}}_T})^{\bf{T}}})\\
  &\hspace*{0.3cm}- 2Tr(\varphi ({{\bf{X}}_{GT}}){\bf{W}}(\varphi {({{\bf{X}}_T})^{\bf{T}}})\\
\end{aligned}
\label{fun501}
\end{equation}
where ${\bf{W}} \in {R^{{n_T} \times {n_T}}}$ is the affinity matrix
described as
${{{W}}_{pq}} = \left\{ {\begin{array}{*{20}{c}}
{1,  if\ {x_{GT}^p} \in {NN_k}({x_T^q}) or \ {x_T^q} \in {NN_k}({x_{GT}^p})}\\
{0,  \ otherwise}
\end{array}} \right.$
and ${NN_k}({\bf{x}})$ represents the $k^{th}$ nearest neighbors of sample ${\bf{x}}$.
The matrix ${\bf{D}} \in {R^{{n_T} \times {n_T}}}$ is a diagonal matrix with entries
${{{D}}_{pp}} = \sum\limits_q {{{{W}}_{pq}}} $, $p=1,...,n_T$.
As claimed before, ${\bm{\mathcal {P}}^{\bf{T}}}{\bf{ = }}{{\bf{\Phi }}^{\bf{T}}}\varphi {{\bf{(X)}}^{\bf{T}}}$, the projected source data and target data can be expressed as ${{\bf{\Phi }}^{\bf{T}}}\varphi {{\bf{(X)}}^{\bf{T}}}\varphi {\bf{(}}{{\bf{X}}_S}{\bf{)}}$ and  ${{\bf{\Phi }}^{\bf{T}}}\varphi {{\bf{(X)}}^{\bf{T}}}\varphi {\bf{(}}{{\bf{X}}_T}{\bf{)}}$. By substituting $\varphi {\bf{(}}{{\bf{X}}_{GT}}{\bf{)}}=\varphi {\bf{(}}{{\bf{X}}_S}{\bf{){\bm{\mathcal{Z}}}}}$ and the Gram matrix after projection (i.e. ${{\bf{\Phi }}^{\bf{T}}}{{\bf{K}}_S}$ and ${{\bf{\Phi }}^{\bf{T}}}{{\bf{K}}_T}$ ) into Eq. (\ref{fun501}), the MC based LGDM loss can be further formulated as
\begin{small}
\begin{equation}
\begin{aligned}
&\mathop {\mathop {\min }\limits_{{\bf{\Phi }},{\bm{\mathcal{Z}}}} \frac{1}{{{{({n_T})}^2}}}}\limits_{} Tr({{\bf{\Phi }}^{\bf{T}}}{{\bf{K}}_S}{\bm{\mathcal{Z}}}{\bf{D}}{({{\bf{\Phi }}^{\bf{T}}}{{\bf{K}}_S}{\bm{\mathcal{Z}}})^{\bf{T}}}) \\
&\hspace*{0.3cm}+\frac{1}{{{{({n_T})}^2}}} Tr({{\bf{\Phi }}^{\bf{T}}}{{\bf{K}}_T}{\bf{D}}{({{\bf{\Phi }}^{\bf{T}}}{{\bf{K}}_T})^{\bf{T}}})\\
&\hspace*{0.3cm}- \frac{2}{{{{({n_T})}^2}}}Tr({{\bf{\Phi }}^{\bf{T}}}{{\bf{K}}_S}{\bm{\mathcal{Z}}}{\bf{W}}{({{\bf{\Phi }}^{\bf{T}}}{{\bf{K}}_T})^{\bf{T}}})
\end{aligned}
\label{fun6}
\end{equation}
\end{small}

From Eq.(\ref{fun6}), the motivation is clearly demonstrated which tends to achieve local structure consistency (i.e. manifold consistency) between the generative target data and the true target data. The intrinsic difference between Eq.(\ref{fun6}) and the manifold embedding or regularization is that we aim to produce the $i.i.d.$ assumption with a manifold criterion, while the conventional manifold learning relies on this assumption.

\subsubsection{Global Generative Discrepancy Metric Loss}
In order to reduce the distribution mismatch between the generative target data and the true target data, a generic MMD  for global generative discrepancy metric (GGDM) is proposed by minimizing the discrepancy as follows.
  \begin{equation}
\begin{aligned}
  GGDM({D_{GT}},{D_T})
   = \frac{1}{{{n_T}}}\sum\limits_{i = 1}^{{n_T}} \left\| {({\varphi {\bf{(X}}_{GT}^{{i}}{\bf{)}} - }  {\varphi {\bf{(X}}_T^{{i}}{\bf{))}}} } \right\|_2^2
\end{aligned}
\end{equation}
where $D_{GT}$ and $D_T$ denote the distribution of generated target domain and true target domain, respectively. However, model may not transfer knowledge directly and it is unclear where a test sample is from ( source or target domain ) if there is not a common subspace. We consider to find a latent common subspace for source and target domain by using a projection matrix $\bm{\mathcal {P}}$. Therefore, by projecting $\varphi ({{\bf{X}}_{GT}})$  and $\varphi ({{\bf{X}}_T})$ to the  subspace, the GGDM loss after projection can be formulated as follows. Considering that $\varphi {\bf{(}}{{\bf{X}}_{GT}}{\bf{)}}=\varphi {\bf{(}}{{\bf{X}}_S}{\bf{){\bm{\mathcal{Z}}}}}$, by substituting it in the equation, there is
  \begin{equation}
\begin{aligned}
GGDM({D_{GT}},{D_T}) &= \frac{1}{{{n_T}}}\sum\limits_{i = 1}^{{n_T}} \left\| {{{\bm{\mathcal {P}}^{\bf{T}}}(\varphi {\bf{(X}}_{GT}^{{i}}{\bf{)}} - } {\varphi {\bf{(X}}_T^{{i}}{\bf{))}}} } \right\|_2^2 \\
&=\frac{1}{{{n_T}}}\left\| {{{\bf{P}}^{\bf{T}}}(\varphi {\bf{(}}{{\bf{X}}_S}{\bf{){\bm{\mathcal{Z}}}}} - \varphi {\bf{(}}{{\bf{X}}_T}){\bf{)1}}} \right\|_2^2
\end{aligned}
\end{equation}
where ${\bf{1}}$ represents a full one column vector.

The projection matrix $\bm{\mathcal {P}}$ is a linear transformation, which can be represented as some linear combination of the training data, i.e. ${\bm{\mathcal {P}}^{\bf{T}}}{\bf{ = }}{{\bf{\Phi }}^{\bf{T}}}\varphi {{\bf{(X)}}^{\bf{T}}}$, where $\bf{\Phi }$ denotes the linear combination coefficient matrix. Then the projected source data can be expressed as ${{\bf{\Phi }}^{\bf{T}}}\varphi {{\bf{(X)}}^{\bf{T}}}\varphi {\bf{(}}{{\bf{X}}_S}{\bf{)}}$ and the projected target data can be expressed as ${{\bf{\Phi }}^{\bf{T}}}\varphi {{\bf{(X)}}^{\bf{T}}}\varphi {\bf{(}}{{\bf{X}}_T}{\bf{)}}$. With the kernel trick, the inner product of implicit transformation is represented as Gram matrix, from raw space to RKHS.
As described in section 4.1, let ${{\bf{K}}_S} = \varphi {{\bf{(X)}}^{\bf{T}}}\varphi ({{\bf{X}}_S})$ and ${{\bf{K}}_T} = \varphi {{\bf{(X)}}^{\bf{T}}}\varphi ({{\bf{X}}_T})$, the source domain and target domain can be expressed simply as ${{\bf{\Phi }}^{\bf{T}}}{{\bf{K}}_S}$ and ${{\bf{\Phi }}^{\bf{T}}}{{\bf{K}}_T}$, respectively. Therefore, the GGDM loss is formulated as
  \begin{equation}
\begin{aligned}
&\mathop {\mathop {\min }\limits_{{\bf{\Phi }},{\bm{\mathcal{Z}}}}}
 \frac{1}{{{n_T}}}\left\| {{{\bf{\Phi }}^{\bf{T}}}({{\bf{K}}_S}{\bm{\mathcal{Z}}}-{{\bf{K}}_T}){\bf{1}}} \right\|_2^2\\
\end{aligned}
\label{fun5}
\end{equation}
\subsubsection{LRC for Domain Correlation Enhancement}
In domain transfer, the loss functions are designed for interpreting the generative target data and the true target data. Significantly, the generative target data plays an critical role in the proposed model. In this work, a general transfer matrix $\bm{\mathcal {Z}}$ is used to bridge the source domain data and the generative data (intermediate result). It is known that for structural consistency between different domains is our goal, therefore, it is natural to consider the low-rank structure of $\bm{\mathcal {Z}}$ as a choice for enhancing the domain correlation. In our MCTL,
low-rank constraint (LRC), that is effective in showing the global structure of different domain data, is finally used. The LRC regularization ensures that the data from different domains can be well interlaced during domain generation, which is significant to reduce the disparity of domain distributions. Furthermore, if the projected data lies in the same manifold, each sample in target domain can be represented by its neighbors in source domain. This requires that the generative transfer matrix $\bm{\mathcal {Z}}$ is approximately block-wise. Therefore, LRC regularization is necessary. Considering the non-convexity property of rank function which is NP-hard, the nuclear norm $||{\bm{\mathcal{Z}}}|{|_*}$ is used as a rank approximation in this work.

\subsubsection{Completed Model of MCTL} By reviewing the MC based LGDM loss in Eq.(\ref{fun6}), the GGDM loss in Eq.(\ref{fun5}), and the LRC regularization, the objective function of our MCTL method is finally formulated as follows.
\begin{small}
  \begin{equation}
\begin{aligned}
&\mathop {\mathop {\min }\limits_{{\bf{\Phi }},{\bm{\mathcal{Z}}}} }\limits_{} \frac{1}{{{{({n_T})}^2}}}Tr({{\bf{\Phi }}^{\bf{T}}}{{\bf{K}}_S}{\bm{\mathcal{Z}}}{\bf{D}}{({{\bf{\Phi }}^{\bf{T}}}{{\bf{K}}_S}{\bm{\mathcal{Z}}})^{\bf{T}}}) \\
&\hspace*{0.3cm}+\frac{1}{{{{({n_T})}^2}}} Tr({{\bf{\Phi }}^{\bf{T}}}{{\bf{K}}_T}{\bf{D}}{({{\bf{\Phi }}^{\bf{T}}}{{\bf{K}}_T})^{\bf{T}}}) \\
&\hspace*{0.3cm}- \frac{2}{{{{({n_T})}^2}}}Tr({{\bf{\Phi }}^{\bf{T}}}{{\bf{K}}_S}{\bm{\mathcal{Z}}}{\bf{W}}{({{\bf{\Phi }}^{\bf{T}}}{{\bf{K}}_T})^{\bf{T}}})\\
&\hspace*{0.3cm}+ \tau \frac{1}{{{n_T}}}\left\| {{{\bf{\Phi }}^{\bf{T}}}({{\bf{K}}_S}{\bm{\mathcal{Z}}}-{{\bf{K}}_T}){\bf{1}}} \right\|_2^2 \\
&\hspace*{0.3cm} + {\lambda _1}||{\bm{\mathcal{Z}}}|{|_*}\\
&s.t.{{\bf{\Phi }}^{\bf{T}}}{\bf{K\Phi }} = {\bf{I}}
\end{aligned}
\label{fun7}
\end{equation}
\end{small}
where $\tau$ and ${\lambda _1}$ are the trade-off parameters. The rows of  $\bm{\mathcal {P}}$ are required to be orthogonal and normalized to unit norm for preventing trivial solutions by enforcing ${\bm{\mathcal {P}}^{\bf{T}}}\bm{\mathcal {P}} = {\bf{I}}$, which can be further rewritten as ${{\bf{\Phi }}^{\bf{T}}}{\bf{K\Phi }} = {\bf{I}}$, an equality constraint. Obviously, the model is non-convex with respect to two variables, but can be solved with the variable alternating strategy, and the optimization algorithm is formulated. 

\subsection{Optimization}
There are two variables $\mathbf\Phi$ and $\bm{\mathcal {Z}}$ in the MCTL model (\ref{fun7}), therefore an efficient variable alternating optimization strategy is naturally considered, i.e. one variable is solved while frozen the other one. First, when $\bm{\mathcal {Z}}$ is fixed, a general Eigen-value decomposition is used for solving $\mathbf\Phi$. Second, when $\mathbf\Phi$ is fixed, the inexact augmented Lagrangian multiplier (IALM) and gradient descent are used to solve $\bm{\mathcal {Z}}$. In the following, the optimization details of the proposed method are presented.

By introducing an auxiliary variable $\bm{\mathcal {J}}$, the problem (\ref{fun7}) can be written as follows. Furthermore, with the augmented Lagrange function\cite{Lin2011Linearized}, the model can be written as
\begin{small}
\begin{equation}
\begin{aligned}
&\mathop {\mathop {\min }\limits_{{\bf{\Phi }},{\bm{\mathcal{Z}}},{\bm{\mathcal {J}}}} }\limits_{} \frac{1}{{{{({n_T})}^2}}}(Tr({{\bf{\Phi }}^{\bf{T}}}{{\bf{K}}_S}{\bm{\mathcal{Z}}}{\bf{D}}{({{\bf{\Phi }}^{\bf{T}}}{{\bf{K}}_S}{\bm{\mathcal{Z}}})^{\bf{T}}})\\
 &\hspace*{0.1cm}+ Tr({{\bf{\Phi }}^{\bf{T}}}{{\bf{K}}_T}{\bf{D}}{({{\bf{\Phi }}^{\bf{T}}}{{\bf{K}}_T})^{\bf{T}}})
- 2Tr({{\bf{\Phi }}^{\bf{T}}}{{\bf{K}}_S}{\bm{\mathcal{Z}}}{\bf{W}}{({{\bf{\Phi }}^{\bf{T}}}{{\bf{K}}_T})^{\bf{T}}}))\\
&\hspace*{0.1cm}+\frac{\tau }{{{{({n_T})}^2}}}{{\bf{\Phi }}^{\bf{T}}}({{\bf{K}}_S}{\bm{\mathcal{Z}}}{\bf{1}}{({{\bf{K}}_S}{\bm{\mathcal{Z}}})^{\bf{T}}} - {{\bf{K}}_S}{\bm{\mathcal{Z}}}{\bf{1}}{({{\bf{K}}_T})^{\bf{T}}}\\
 &\hspace*{0.1cm}- {{\bf{K}}_T}{\bf{1}}{{\bm{\mathcal{Z}}}^{\bf{T}}}{({{\bf{K}}_S})^{\bf{T}}} + {{\bf{K}}_T}{\bf{1}}{({{\bf{K}}_T})^{\bf{T}}}){\bf{\Phi }} + {\lambda _1}||{\bm{\mathcal {J}}}|{|_*}\\
 &\hspace*{0.1cm}+ Tr({\bm{\mathcal {R}}}_{\bf{1}}^{\bf{T}}{\bf{({\bm{\mathcal{Z}}} - \bm{\mathcal {J}})}}) + \frac{\mu }{2}{\rm{(||}}{\bf{{\bm{\mathcal{Z}}} - \bm{\mathcal {J}}}}||_F^2)
\end{aligned}
\label{fun9}
\end{equation}
\end{small}
where {\bf{1}} represents a full one matrix instead of a full one vector as the problem (\ref{fun7}) is unfolded. $\bm{\mathcal {R}}_1$ denotes the Lag-multiplier and $\mu$ is a penalty parameter.

In the following, we present how to optimize the three variables $\mathbf\Phi$, $\bm{\mathcal {J}}$, and $\bm{\mathcal{Z}}$ in the problem (\ref{fun9}) based on Eigen-value decomposition, IALM and gradient descent in step-wise.

\subsubsection{Update $\mathbf\Phi$}
By frozen $\bm{\mathcal{Z}}$ and $\bm{\mathcal {J}}$, $\mathbf\Phi$ can be solved as
\begin{small}
\begin{equation}
\begin{aligned}
&{{\bf{\Phi }}^ * } = {\rm{arg }}\mathop {\mathop {\min }\limits_{\bf{\Phi }} }\limits_{} \frac{1}{{{{({n_T})}^2}}}(Tr({{\bf{\Phi }}^{\bf{T}}}{{\bf{K}}_S}{\bm{\mathcal{Z}}}{\bf{D}}{({{\bf{\Phi }}^{\bf{T}}}{{\bf{K}}_S}{\bm{\mathcal{Z}}})^{\bf{T}}}) \\
&\hspace*{0.1cm}+ Tr({{\bf{\Phi }}^{\bf{T}}}{{\bf{K}}_T}{\bf{D}}{({{\bf{\Phi }}^{\bf{T}}}{{\bf{K}}_T})^{\bf{T}}}) - 2Tr({{\bf{\Phi }}^{\bf{T}}}{{\bf{K}}_S}{\bm{\mathcal{Z}}}{\bf{W}}{({{\bf{\Phi }}^{\bf{T}}}{{\bf{K}}_T})^{\bf{T}}}))\\
&\hspace*{0.1cm} + \frac{\tau }{{{{({n_T})}^2}}}{{\bf{\Phi }}^{\bf{T}}}({{\bf{K}}_S}{\bm{\mathcal{Z}}}{\bf{1}}{{\bm{\mathcal{Z}}}^{\bf{T}}}{({{\bf{K}}_S})^{\bf{T}}} - {{\bf{K}}_S}{\bm{\mathcal{Z}}}{\bf{1}}{({{\bf{K}}_T})^{\bf{T}}}\\
 &\hspace*{0.1cm}- {{\bf{K}}_T}{\bf{1}}{{\bm{\mathcal{Z}}}^{\bf{T}}}{({{\bf{K}}_S})^{\bf{T}}} + {{\bf{K}}_T}{\bf{1}}{({{\bf{K}}_T})^{\bf{T}}}){\bf{\Phi }}\\
 &s.t.{{\bf{\Phi }}^{\bf{T}}}{\bf{K\Phi }} = {\bf{I}}\\
\end{aligned}
\label{fun10}
\end{equation}
\end{small}

We can derive the solution $\mathbf\Phi_{K}$ of the $K^{th}$ iteration in column-wise. To obtain the $i^{th}$ column vector in $\mathbf\Phi_{K}$, by setting the partial derivative of problem (\ref{fun10}) with respect to $\mathbf\Phi_{K(:,i)}$ to be zero, there is
\begin{small}
\begin{equation}
\begin{aligned}
&\frac{1}{{{{({n_T})}^2}}}({{\bf{K}}_S}{\bm{\mathcal{Z}}}{\bf{D}}{{\bm{\mathcal{Z}}}^{\bf{T}}}{({{\bf{K}}_S})^{\bf{T}}} + {{\bf{K}}_T}{\bf{D}}{({{\bf{K}}_T})^{\bf{T}}} - {{\bf{K}}_S}{\bm{\mathcal{Z}}}{\bf{W}}{({{\bf{K}}_T})^{\bf{T}}} \\
&- {{\bf{K}}_T}{\bf{W}}{{\bm{\mathcal{Z}}}^{\bf{T}}}{({{\bf{K}}_S})^{\bf{T}}}){{\bf{\Phi }}_{K(:,i)}} + \frac{\tau }{{{{({n_T})}^2}}}({{\bf{K}}_S}{\bm{\mathcal{Z}}}1{{\bm{\mathcal{Z}}}^{\bf{T}}}{({{\bf{K}}_S})^{\bf{T}}}\\
&- {{\bf{K}}_S}{\bm{\mathcal{Z}}}{\bf{1}}{({{\bf{K}}_T})^{\bf{T}}} - {{\bf{K}}_T}{\bf{1}}{{\bm{\mathcal{Z}}}^{\bf{T}}}{({{\bf{K}}_S})^{\bf{T}}} + {{\bf{K}}_T}{\bf{1}}{({{\bf{K}}_T})^{\bf{T}}}){{\bf{\Phi }}_{K(:,i)}}\\
 &=  - \lambda {\bf{K}}{{\bf{\Phi }}_{K(:,i)}}
\end{aligned}
\label{fun11}
\end{equation}
\end{small}
It is clear that $\mathbf\Phi_{K}$ can be obtained by solving an Eigen-decomposition problem, and $\mathbf\Phi_{K(:,i)}$ is the $i^{th}$ eigenvector corresponding to the $i^{th}$ smallest eigenvalue.

\subsubsection{Update ${\mathcal{J}}$}
By frozen $\mathbf\Phi$ and $\bm{\mathcal{Z}}$, the problem is solved with respect to $\bm{\mathcal{J}}$.
After dropping out the irrelevant terms with respect to $\bm{\mathcal{J}}$, $\bm{\mathcal{J}_{K+1}}$ in iteration $K+1$ can be solved as
\begin{small}
\begin{equation}
\begin{aligned}
\bm{\mathcal{J}}_{K+1}=&\min_{\bm{\mathcal{J}}_K}\lambda_1\parallel\bm{\mathcal{J}}_{K}\parallel_*
+\mathbf{Tr}(\bm{\mathcal{R}}_{1K}^T(\bm{\mathcal{Z}}_{K}-\bm{\mathcal{J}}_{K}))\\
&+\frac{\mu_K}2\parallel\bm{\mathcal{Z}}_{K}-\bm{\mathcal{J}}_{K}\parallel_F^2
\end{aligned}
\label{fun12}
\end{equation}
\end{small}

It can be further rewritten as
\begin{small}
\begin{equation}
\begin{aligned}
\bm{\mathcal{J}}_{K+1}=\min_{\bm{\mathcal{J}}_K}\lambda_1\parallel\bm{\mathcal{J}}_{K}\parallel_*
+\frac{\mu_K}2\parallel\bm{\mathcal{J}}_{K}-(\bm{\mathcal{Z}}_{K}+\frac{\bm{\mathcal{R}}_{1K}}{\mu_K})\parallel_F^2
\end{aligned}
\label{fun13}
\end{equation}
\end{small}

 Problem (\ref{fun13}) can be efficiently solved using the singular value thresholding (SVT) operator\cite{Cai2008A}, which contains two major steps. First, singular value decomposition (SVD) is conducted on matrix $\bm{\mathcal{S}} = \bm{\mathcal{Z}}_{K}+\frac{\bm{\mathcal{R}}_{1K}}{\mu_K}$, and get $\bm{\mathcal{S}} = \mathbf{U_S}\mathbf{\sum_S}\mathbf{V_S}$, where $\mathbf{\sum_S} = diag(\{{\sigma_i}\}_{1\le i\le r})$, ${\sigma_i}$ is the singular value with rank $r$. Second, the optimal solution $\bm{\mathcal{J}}_{K+1}$ is then obtained by thresholding the singular values as $\bm{\mathcal{J}}_{K+1}=\mathbf{U_S}\mathbf{\Omega_{(1/{\mu_k})}}(\mathbf{\sum_S})\mathbf{V_S}$, where $\mathbf{\Omega_{(1/{\mu_k})}}(\mathbf{\sum_S})=diag(\{{\sigma_i}-(1/{\mu_k})\}_+)$, and $\{\bullet\}_+$ denotes the positive value operator.

\subsubsection{Update ${\mathcal{Z}}$}
By frozen $\mathbf\Phi$ and $\bm{\mathcal{J}}$, the problem is solved with respect to $\bm{\mathcal{Z}}$.
By dropping out those terms independent of $\bm{\mathcal{Z}}$ in (\ref{fun9}), there is
\begin{small}
\begin{equation}
\begin{aligned}
&\min_{\bm{\mathcal{Z}}}\frac{1}{{{{({n_T})}^2}}}(Tr({{\bf{\Phi }}^{\bf{T}}}{{\bf{K}}_S}{\bm{\mathcal{Z}}}{\bf{D}}{({{\bf{\Phi }}^{\bf{T}}}{{\bf{K}}_S}{\bm{\mathcal{Z}}})^{\bf{T}}})\\
 &\hspace*{0.5cm}- 2Tr({{\bf{\Phi }}^{\bf{T}}}{{\bf{K}}_S}{\bm{\mathcal{Z}}}{\bf{W}}{({{\bf{\Phi }}^{\bf{T}}}{{\bf{K}}_T})^{\bf{T}}}))+ Tr({\bf{R}}_{\bf{1}}^{\bf{T}}{\bf{({\bm{\mathcal{Z}}} - J)}})\\
&\hspace*{0.5cm}+ \frac{\mu }{2}{\rm{(||}}{\bf{{\bm{\mathcal{Z}}} - J}}||_F^2) + \frac{\tau }{{{{({n_T})}^2}}}{{\bf{\Phi }}^{\bf{T}}}({{\bf{K}}_S}{\bm{\mathcal{Z}}}{\bf{1}}{{\bm{\mathcal{Z}}}^{\bf{T}}}{({{\bf{K}}_S})^{\bf{T}}}\\
 &\hspace*{0.5cm}- {{\bf{K}}_S}{\bm{\mathcal{Z}}}{\bf{1}}{({{\bf{K}}_T})^{\bf{T}}} - {{\bf{K}}_T}{\bf{1}}{\bm{\mathcal{Z}}^{\bf{T}}}{({{\bf{K}}_S})^{\bf{T}}}){\bf{\Phi }}
\end{aligned}
\label{fun14}
\end{equation}
\end{small}

We can see from problem (\ref{fun14}) that it is hard to obtain a closed-form solution of $\bm{\mathcal{Z}}$. Therefore, the general gradient descent operator\cite{Rosasco2009Iterative} is used, and the solution of $\bm{\mathcal{Z}}_{K+1}$ in the $(K+1)^{th}$ iteration is presented as
\begin{equation}
\begin{aligned}
\bm{\mathcal{Z}}_{K+1}=\bm{\mathcal{Z}}_{K}-\alpha\bullet\bigtriangledown(\bm{\mathcal{Z}})
\end{aligned}
\label{fun15}
\end{equation}
where $\bigtriangledown(\bm{\mathcal{Z}})$ denotes the gradient, which is calculated as
\begin{equation}
\begin{aligned}
&\nabla ({\bm{\mathcal{Z}}}) = \frac{2}{{{{({n_T})}^2}}}({({{\bf{K}}_S})^{\bf{T}}}{\bf{\Phi }}{{\bf{\Phi }}^{\bf{T}}}{{\bf{K}}_S}{\bm{\mathcal{Z}}}{\bf{D}} - {({{\bf{K}}_S})^{\bf{T}}}{\bf{\Phi }}{{\bf{\Phi }}^{\bf{T}}}{{\bf{K}}_T}{\bf{W}})\\
&\hspace*{1.0cm} + {{\bf{R}}_1} + \mu {\rm{(}}{\bf{{\bm{\mathcal{Z}}} - J}})+ \frac{{2\tau }}{{{{({n_T})}^2}}}{({{\bf{K}}_S})^{\bf{T}}}{\bf{\Phi }}{{\bf{\Phi }}^{\bf{T}}}{{\bf{K}}_S}{\bm{\mathcal{Z}}}{\bf{1}}\\
&\hspace*{1.0cm} - \frac{{2\tau }}{{{{({n_T})}^2}}}{({{\bf{K}}_S})^{\bf{T}}}{\bf{\Phi }}{{\bf{\Phi }}^{\bf{T}}}{{\bf{K}}_T}{\bf{1}}
\end{aligned}
\label{fun16}
\end{equation}

In detail, the iterative optimization procedure of the proposed MCTL is summarized in \textbf{Algorithm 1}.

\begin{table}[t]
\begin{tabular}{l}
\toprule
\textbf{Algorithm 1}  The Proposed MCTL\\
\midrule
\textbf{Input:} $\bm{\mathcal{X}}_{S}\in\mathcal{R}^{m\times n_{S}}$, $\bm{\mathcal{X}}_{T}\in\mathcal{R}^{m\times n_{T}}$, $\tau$, $\lambda_1$ \\
\textbf{Procedure:}\\
1. Compute $\bm{\mathcal{K}}_T=\varphi(\bm{\mathcal{X}})^T\varphi(\bm{\mathcal{X}}_T)$, $\bm{\mathcal{K}}_S=\varphi(\bm{\mathcal{X}})^T\varphi(\bm{\mathcal{X}}_S)$,\\
\hspace*{1.4cm} $\bm{\mathcal{K}}=\varphi(\bm{\mathcal{X}})^T\varphi(\bm{\mathcal{X}})$, $\bm{\mathcal{X}}=[\bm{\mathcal{X}}_S, \bm{\mathcal{X}}_T]$\\
2.Initialize: $\bm{\mathcal{J}}$=$\bm{\mathcal{Z}}$=$\bf{0}$\\
3.	\textbf{While} not converge \textbf{do}\\
\hspace*{0.5cm}3.1	\textbf{Step1}: Fix $\bm{\mathcal{J}}$ and $\bm{\mathcal{Z}}$, and update $\mathbf{\Phi}$ by solving\\
\hspace*{1cm}eigenvalue decomposition problem (\ref{fun11}).\\
\hspace*{0.5cm}3.2	\textbf{Step2}: Fix $\mathbf{\Phi}$, and update $\bm{\mathcal{Z}}$ using IALM: \\
\hspace*{1cm} 3.2.1. Fix $\bm{\mathcal{Z}}$ and update $\bm{\mathcal{J}}$ by using the singular value \\
\hspace*{1cm} thresholding (SVT) \cite{Cai2008A} operator on problem (\ref{fun13}).\\
\hspace*{1cm} 3.2.2. Fix $\bm{\mathcal{J}}$ and update $\bm{\mathcal{Z}}$ according to gradient\\
\hspace*{1cm} descent operator, i.e. Equation (\ref{fun15}).\\
\hspace*{0.5cm}3.3	Update the multiplier $\bm{\mathcal{R}}_1$:\\
\hspace*{1.5cm} $\bm{\mathcal{R}}_1=\bm{\mathcal{R}}_1+\mu(\bm{\mathcal{Z}}-\bm{\mathcal{J}})$\\
\hspace*{0.5cm}3.4	Update the parameter $\mu$:\\
\hspace*{1.5cm} $\mu=min(\mu \times 1.01, max_{\mu})$\\
\hspace*{0.5cm}3.5	Check convergence\\
\textbf{end while}\\
\textbf{Output:} $\mathbf\Phi$ and $\bm{\mathcal{Z}}$.\\
\bottomrule
\end{tabular}
\label{Tab}
\end{table}

\section{MCTL-S: Simplified Version of MCTL}
As illustrated in MCTL, which aims to minimize the distribution discrepancy between the generative target data and the true target data as close as possible, by using the manifold criterion. In this section, considering the generic manifold embedding, for model simplicity, we rewrite a simplified version of MCTL (MCTL-S in short) as illustrated in Fig.\ref{figAMCTL}.
\subsection{Formulation of MCTL-S}
\begin{figure}[t]
\centering
  \includegraphics[width=1\linewidth]{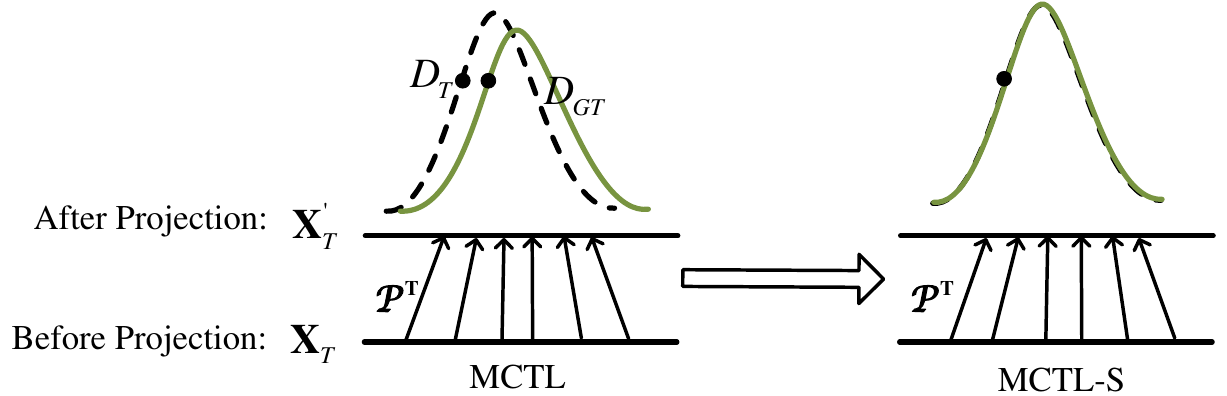}
   \caption{Difference between MCTL (left) and MCTL-S (right). In MCTL, there is error between the true target domain $D_T$ and the generative target domain $D_{GT}$. In MCTL-S, the $D_{GT}$ is supposed to be coincided with the true target domain $D_T$.}
   \label{figAMCTL}
\end{figure}

With the description of Fig.\ref{figAMCTL} (right), suppose an extreme case of $perfect$ domain generation, that is, the generated target data is strictly the same as the true target data, i.e. ${\textbf{X}_{GT}}={\textbf{X}_T}$ ($D_{GT}$ coincides with $D_{T}$), then MCTL-S is formulated as,

\begin{equation}
\begin{aligned}
&\mathop {\min }\limits_{{\bf{\Phi }},\bm{\mathcal{Z}}} \frac{2}{{{{({n_T})}^2}}}Tr({{\bf{\Phi }}^{\bf{T}}}{{\bf{K}}_S}{\bf{\bm{\mathcal{Z}}L}}{({{\bf{\Phi }}^{\bf{T}}}{{\bf{K}}_S}\bm{\mathcal{Z}})^{\bf{T}}})\\
&\hspace*{0.5cm} + \tau \left\| {\frac{1}{{{n_T}}}{{\bf{\Phi }}^{\bf{T}}}({{\bf{K}}_S}{\bm{\mathcal{Z}}}-{{\bf{K}}_T}){\bf{1}}} \right\|_2^2 \\
&\hspace*{0.5cm}+ {\lambda _1}||\bm{\mathcal{Z}}|{|_*}
\end{aligned}
\label{fun17}
\end{equation}

where ${\bf{L}}{\rm{ = }}{\bf{D - W}}$ is the conventional Laplacian matrix. Also, the objective function (\ref{fun17}) contains three items such as the MC based LGDM loss, the GGDM loss and LRC regularization. From the MC-S loss term in Equation (\ref{fun17}), we observe a generic manifold regularization term with Laplacian matrix. Therefore, the MC loss can be degenerated into a conventional manifold constraint by implying ${{\bf{\Phi }}^{\bf{T}}}{{\bf{K}}_T}{\rm{ = }}{{\bf{\Phi }}^{\bf{T}}}{{\bf{K}}_S}\bm{\mathcal{Z}}$, which shows that MCTL-S model is harsher than MCTL model.

The following experimental results in Table \ref{tab12} and \ref{tab13} also prove that both the harsh MCTL-S model and the MCTL can achieve good performance. This demonstrates that manifold criterion based intermediate domain generation is a very effective scheme for transfer learning.

\subsection{Optimization of MCTL-S}
MCTL-S has a similar mechanism with MCTL, therefore, the MCTL-S optimization is almost the same as MCTL. With two updating steps for $\mathbf\Phi$ and $\bm{\mathcal {Z}}$, the optimization procedure of the MCTL-S method is illustrated as follows.

$\bullet$ Update $\mathbf\Phi$. In the MCTL-S model, by frozen $\bm{\mathcal {Z}}$ and $\bm{\mathcal {J}}$, the derivative of the objective function (\ref{fun17}) w.r.t. $\mathbf\Phi_{K(:,i)}$ is set as zero, there is
\begin{equation}
\begin{aligned}
&\frac{2}{{{{({n_T})}^2}}}({{\bf{K}}_S}{\bm{\mathcal{Z}}}{\bf{L}}{{\bm{\mathcal{Z}}}^{\bf{T}}}{({{\bf{K}}_S})^{\bf{T}}}){{\bf{\Phi }}_{K(:,i)}} 
+ \frac{\tau }{{{{({n_T})}^2}}}({{\bf{K}}_S}{\bm{\mathcal{Z}}}1{{\bm{\mathcal{Z}}}^{\bf{T}}}{({{\bf{K}}_S})^{\bf{T}}}\\
&- {{\bf{K}}_S}{\bm{\mathcal{Z}}}1{({{\bf{K}}_T})^{\bf{T}}} - {{\bf{K}}_T}1{{\bm{\mathcal{Z}}}^{\bf{T}}}{({{\bf{K}}_S})^{\bf{T}}} + {{\bf{K}}_T}1{({{\bf{K}}_T})^{\bf{T}}}){{\bf{\Phi }}_{K(:,i)}}\\
 &=  - \lambda {\bf{K}}{{\bf{\Phi }}_{K(:,i)}}
\end{aligned}
\label{fun18}
\end{equation}

Therefore, $\mathbf\Phi_{K}$ in iteration $K$ can be obtained by solving an Eigenvalue decomposition problem, and $\mathbf\Phi_{K(:,i)}$ is the $i^{th}$ eigenvector corresponding to the $i^{th}$ smallest eigenvalue.

$\bullet$ Update $\bm{\mathcal{J}}$. The variable $\bm{\mathcal{J}}$ can be effectively solved by the singular value thresholding (SVT) operator\cite{Cai2008A}, which is similar to the problem (\ref{fun13}).

$\bullet$ Update $\bm{\mathcal{Z}}$. The variable $\bm{\mathcal {Z}}$ can be updated according to section 4.3.3 by using gradient descent algorithm. The gradient with respect to $\bm{\mathcal{Z}}$ can be expressed as
\begin{equation}
\begin{aligned}
&\nabla ({\bm{\mathcal{Z}}}) = \frac{4}{{{{({n_T})}^2}}}{({{\bf{K}}_S})^{\bf{T}}}{\bf{\Phi }}{{\bf{\Phi }}^{\bf{T}}}{{\bf{K}}_S}{\bm{\mathcal{Z}}}{\bf{L}} 
 + {{\bf{R}}_1} + \mu {\rm{(}}{\bf{{\bm{\mathcal{Z}}} - J}})\\
&\hspace*{0.0cm}+ \frac{{2\tau }}{{{{({n_T})}^2}}}({({{\bf{K}}_S})^{\bf{T}}}{\bf{\Phi }}{{\bf{\Phi }}^{\bf{T}}}{{\bf{K}}_S}{\bm{\mathcal{Z}}}1
 -{({{\bf{K}}_S})^{\bf{T}}}{\bf{\Phi }}{{\bf{\Phi }}^{\bf{T}}}{{\bf{K}}_T}1)
\end{aligned}
\label{fun19}
\end{equation}

\section{Classification}
For classification, the projected source data and target data can be represented as ${\bm{\mathcal {X}}_S}'=\mathbf\Phi^{\bf{T}}\varphi(\bm{\mathcal {X}})^{\bf{T}}\varphi(\bm{\mathcal {X}}_S)$, ${\bm{\mathcal {X}}_T}'=\mathbf\Phi^{\bf{T}}\varphi(\bm{\mathcal {X}})^{\bf{T}}\varphi(\bm{\mathcal {X}}_S){\bm{\mathcal{Z}}}$. Then, existing classifiers (e.g., SVM, least square method\cite{kanamori2009least}, SRC\cite{wright2009robust}) can be trained on the domain aligned and augmented training data $[{\bm{\mathcal {X}}_S}',{\bm{\mathcal {X}}_T}']$ with label $\bm{\mathcal {Y}}=[{\bm{\mathcal {Y}}_S},{\bm{\mathcal {Y}}_T}]$ by following the experimental setting as LSDT\cite{zhang2016lsdt}. Notably, for the COIL-20, MSRC and VOC2007 experiments, in order to follow the same experimental setting with DTSL\cite{Xu2015}, the classifier is trained only on ${\bm{\mathcal {X}}_S}'$ with label ${\bm{\mathcal {Y}}_S}$. Finally, classification on those unlabeled target test data, i.e. ${\bm{\mathcal {X}}_{Tu}}'=\mathbf\Phi^{\bf{T}}\varphi(\bm{\mathcal {X}})^{\bf{T}}\varphi(\bm{\mathcal {X}}_{Tu})$, is achieved, and the recognition accuracy is reported and compared.

\begin{figure}
  \begin{minipage}[t]{0.5\linewidth} 
   \centering
    \includegraphics[width=0.3\linewidth,height=1.3cm]{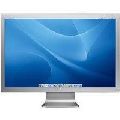}
     \includegraphics[width=0.3\linewidth,height=1.3cm]{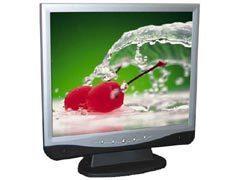}
      \includegraphics[width=0.3\linewidth,height=1.3cm]{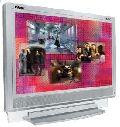}
       \includegraphics[width=0.3\linewidth,height=1.3cm]{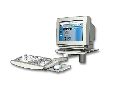}
        \includegraphics[width=0.3\linewidth,height=1.3cm]{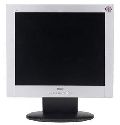}
         \includegraphics[width=0.3\linewidth,height=1.3cm]{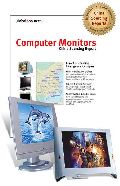}
  \caption*{ Caltech 256 }
   \label{fig:side:a}
  \end{minipage}
  \begin{minipage}[t]{0.49\linewidth}
   \centering
    \includegraphics[width=0.3\linewidth,height=1.3cm]{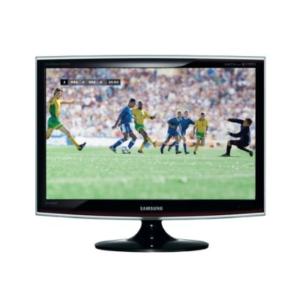}
     \includegraphics[width=0.3\linewidth,height=1.3cm]{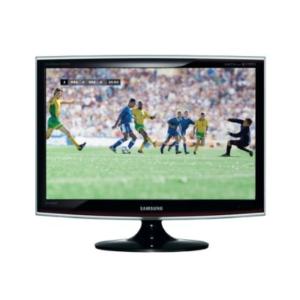}
      \includegraphics[width=0.3\linewidth,height=1.3cm]{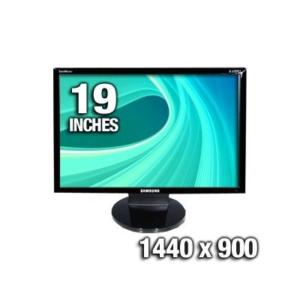}
       \includegraphics[width=0.3\linewidth,height=1.3cm]{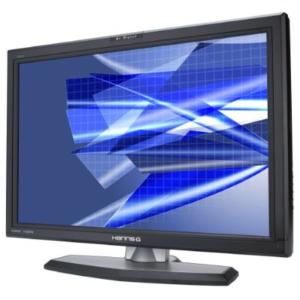}
        \includegraphics[width=0.3\linewidth,height=1.3cm]{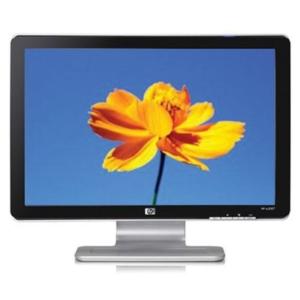}
         \includegraphics[width=0.3\linewidth,height=1.3cm]{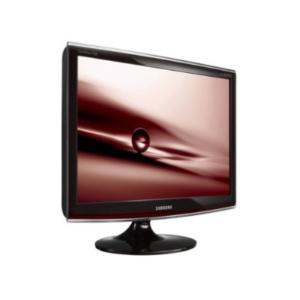}
    \caption*{Amazon}
    \label{fig:side:b}
  \end{minipage}
    \begin{minipage}[t]{0.49\linewidth}
   \centering
    \includegraphics[width=0.3\linewidth,height=1.3cm]{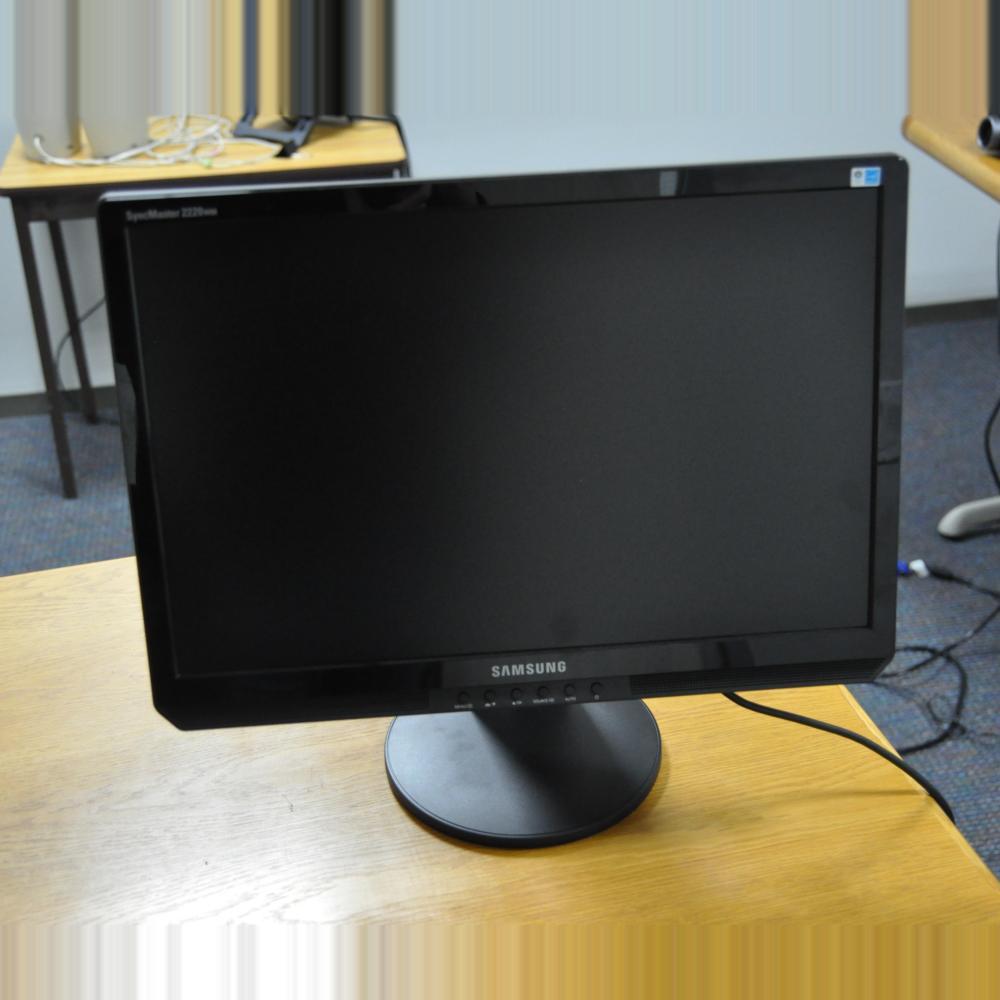}
     \includegraphics[width=0.3\linewidth,height=1.3cm]{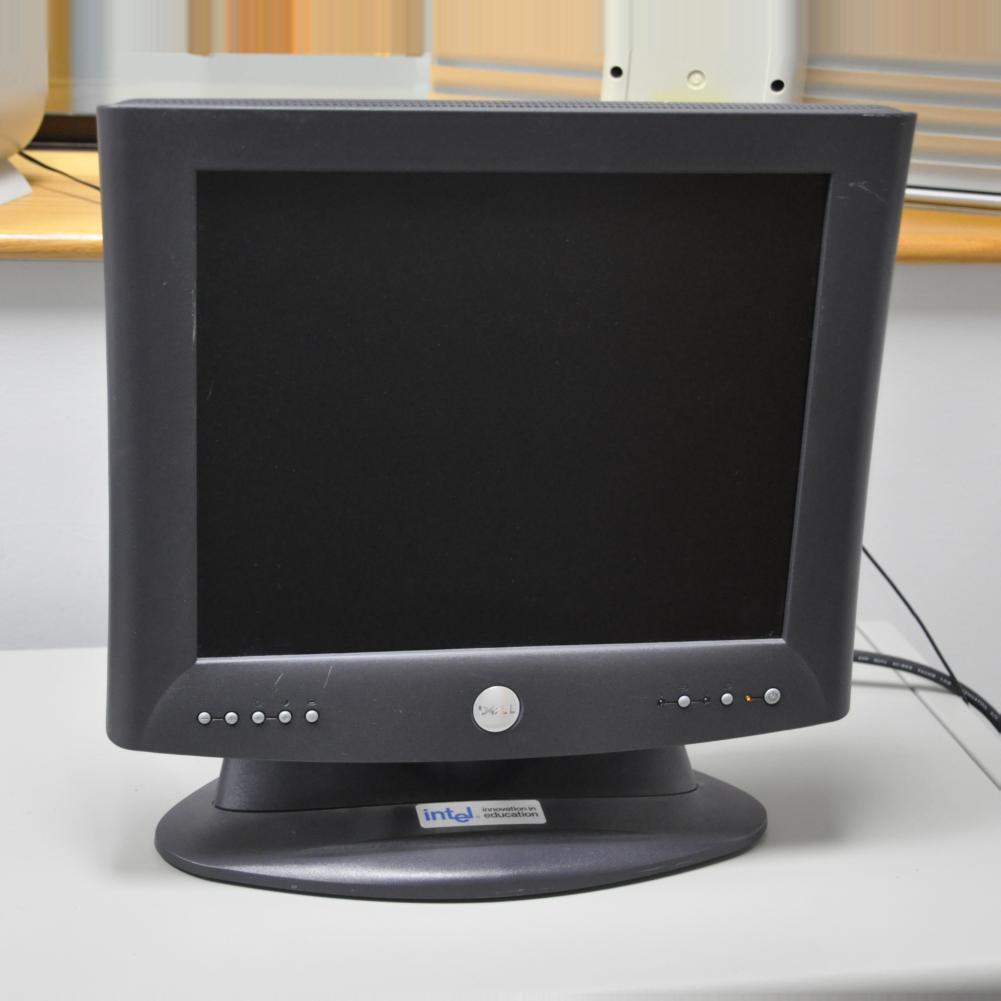}
      \includegraphics[width=0.3\linewidth,height=1.3cm]{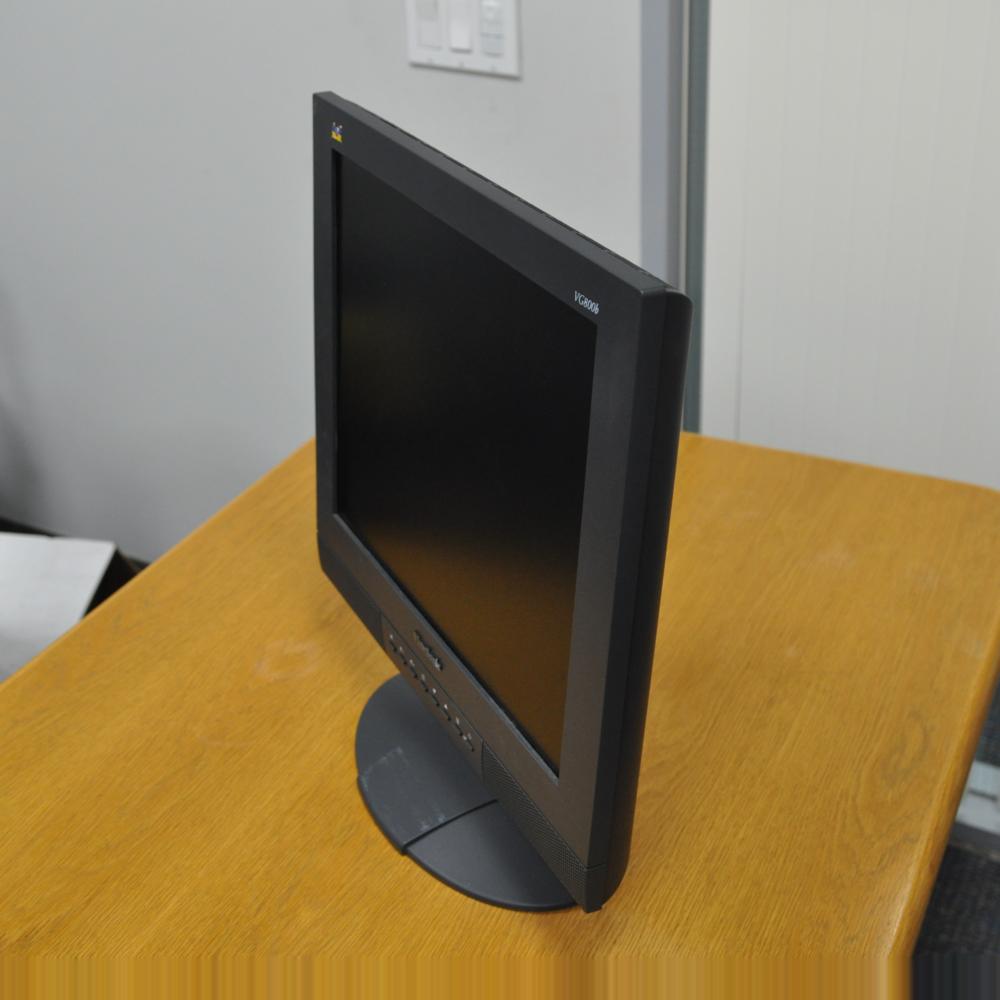}
       \includegraphics[width=0.3\linewidth,height=1.3cm]{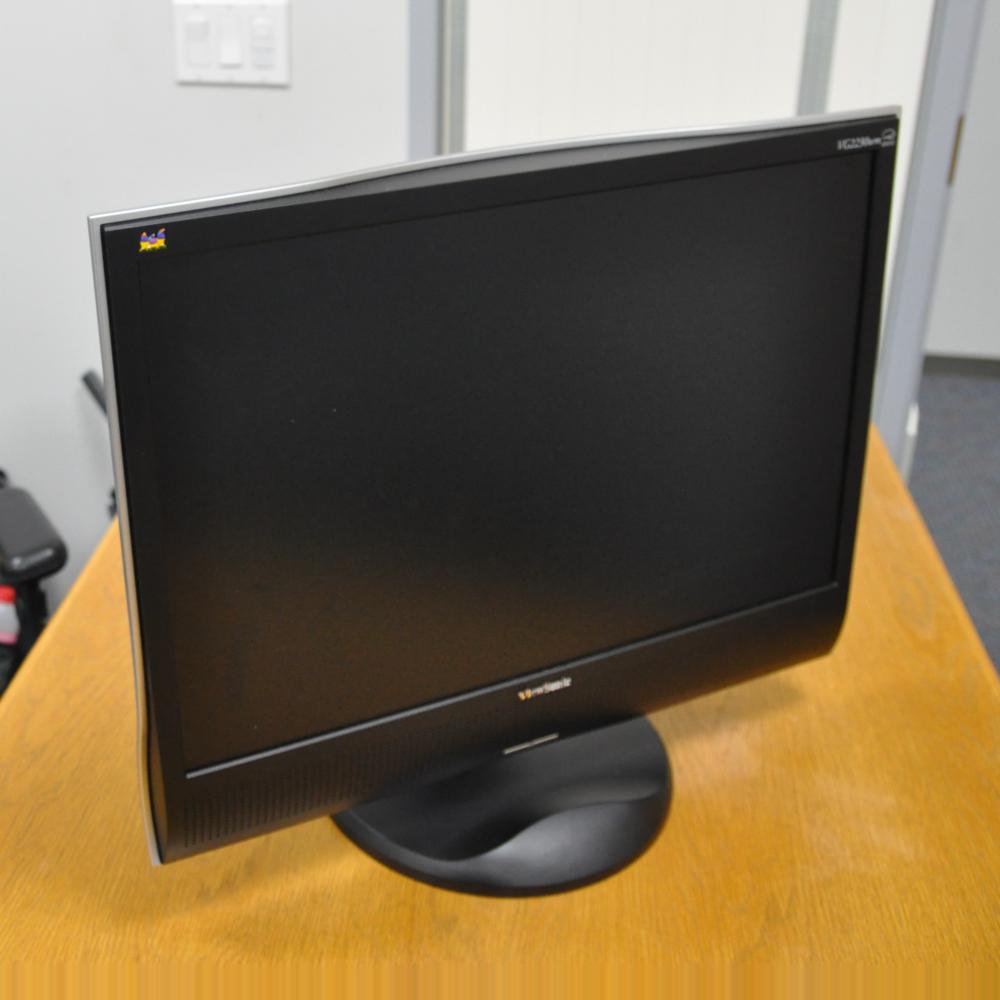}
        \includegraphics[width=0.3\linewidth,height=1.3cm]{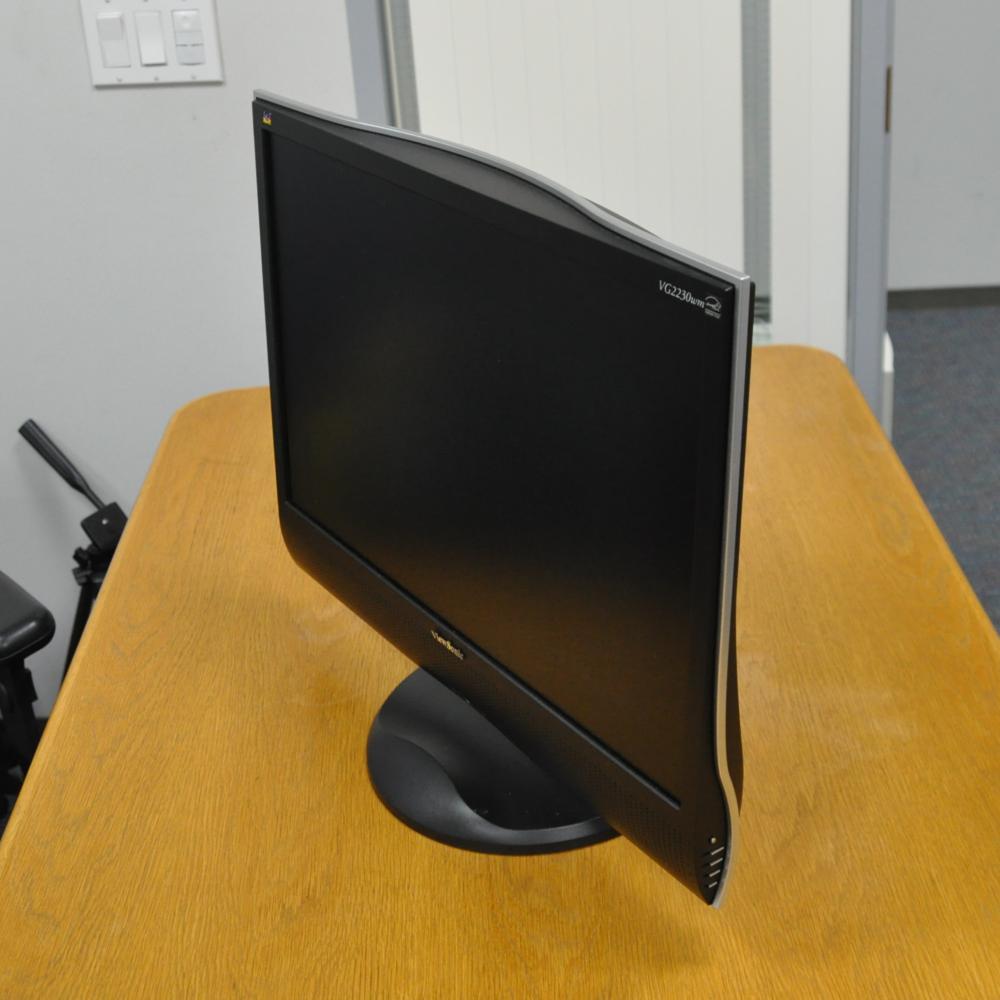}
         \includegraphics[width=0.3\linewidth,height=1.3cm]{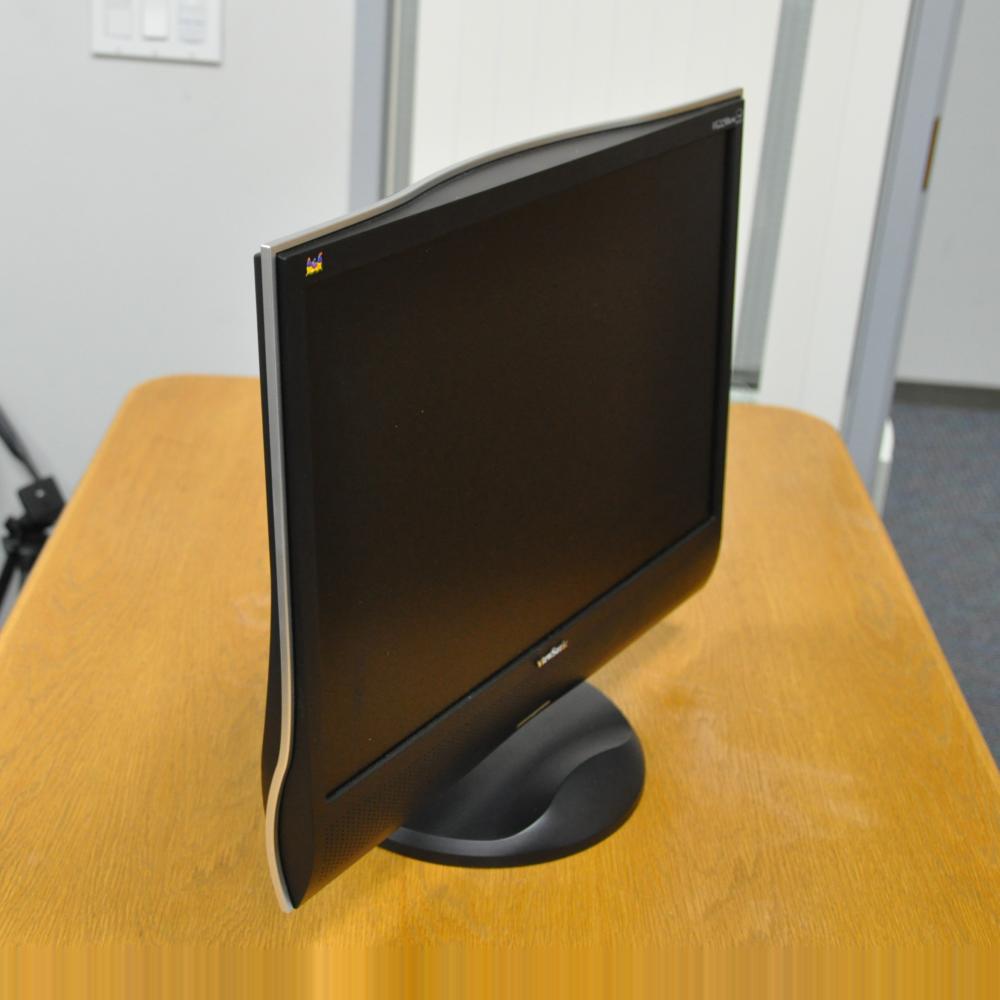}
    \caption*{DSLR}
    \label{fig:side:c}
  \end{minipage}
    \begin{minipage}[t]{0.49\linewidth}
   \centering
    \includegraphics[width=0.3\linewidth,height=1.3cm]{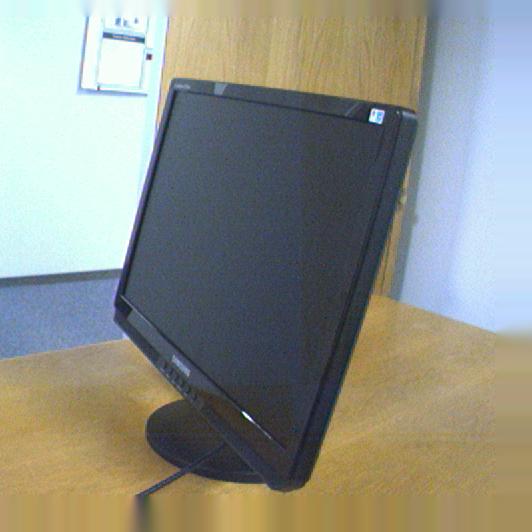}
     \includegraphics[width=0.3\linewidth,height=1.3cm]{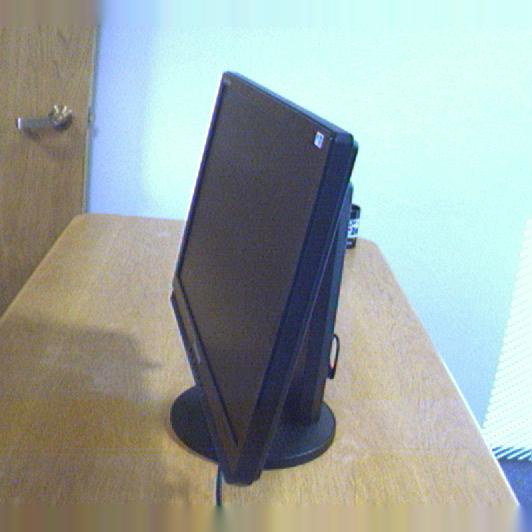}
      \includegraphics[width=0.3\linewidth,height=1.3cm]{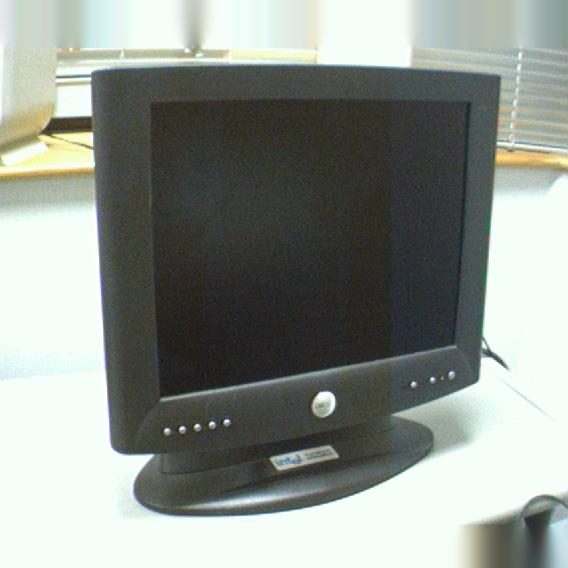}
       \includegraphics[width=0.3\linewidth,height=1.3cm]{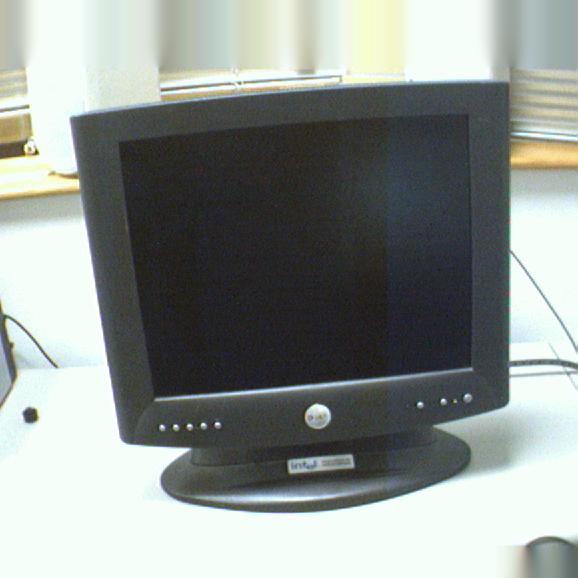}
        \includegraphics[width=0.3\linewidth,height=1.3cm]{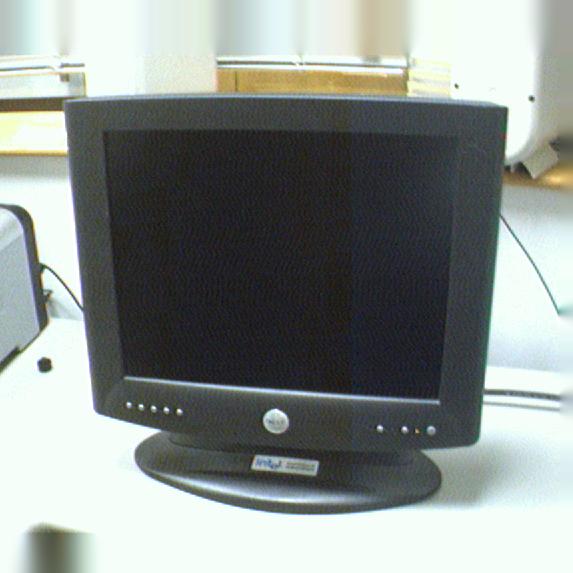}
         \includegraphics[width=0.3\linewidth,height=1.3cm]{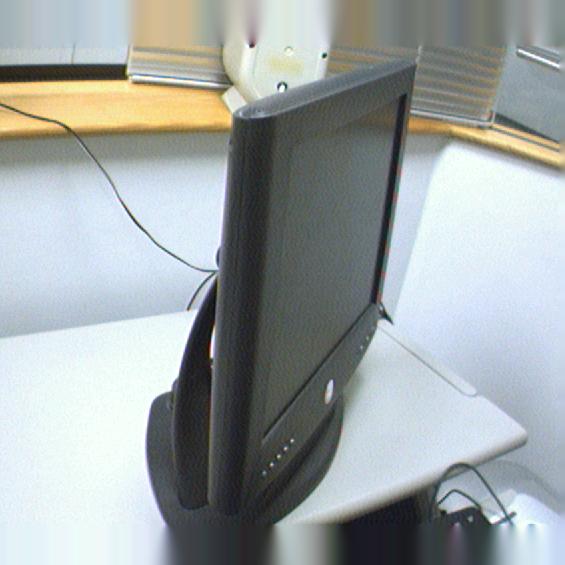}
    \caption*{Webcam}
    \label{fig:side:d}
  \end{minipage}
  \captionsetup{justification=centering}
   \caption{Some images from 4DA datasets}
   \label{fig3}
\end{figure}

\section{Experiments}

In this section, the experiments on several benchmark datasets\cite{gaidon2014self} have been exploited for evaluating the proposed MCTL method, including (1) cross-domain object recognition\cite{gong2014learning},\cite{liu2010hierarchical}: 4DA office data, 4DA-CNN office data, COIL-20 data, and MSRC-VOC 2007 datasets \cite{long2013transfer}; (2) cross-pose face recognition: Multi-PIE face dataset; (3) cross-domain handwritten digit recognition: USPS, SEMEION and MNIST datasets. Several related transfer learning methods based on feature transformation and reconstruction, such as SGF\cite{Gopalan2011Domain}, GFK\cite{Gong2012Geodesic}, SA\cite{Fernando2014Unsupervised}, LTSL\cite{shao2014generalized}, DTSL\cite{Xu2015}, and LSDT\cite{zhang2016lsdt} have been compared and discussed.

\subsection{Cross-domain Object Recognition}
For cross-domain object/image recognition, 5 benchmark datasets are used, where several sample images in 4DA office dataset are shown in Fig. \ref{fig3}, several sample images in COIL-20 object dataset are shown in Fig. \ref{fig4}, several sample images in MSRC and VOC 2007 datasets are described in Fig. \ref{fig5}.

\textbf{Results on 4DA Office dataset (Amazon, DSLR, Webcam\footnote{\url{http://www.eecs.berkeley.edu/~mfritz/domainadaptation/}} and Caltech 256\footnote{\url{http://www.vision.caltech.edu/Image_Datasets/Caltech256/}})}\cite{Gong2012Geodesic}:

Four domains such as Amazon (A), DSLR (D), Webcam (W), and Caltech (C) are included in 4DA dataset, which contains 10 object classes. In our experiment, the configuration is followed in\cite{Gong2012Geodesic} where 20 samples per class are selected from Amazon, 8 samples per class from DSLR, Webcam and Caltech when they are used as source domains; 3 samples per class are chosen when they are used as target training data, while the rest data in target domains are used for testing. Note that the 800-bin SURF features \cite{Gong2012Geodesic},\cite{saenko2010eccv} are extracted.
\begin{figure}[t]
\centering
  \includegraphics[width=1\linewidth]{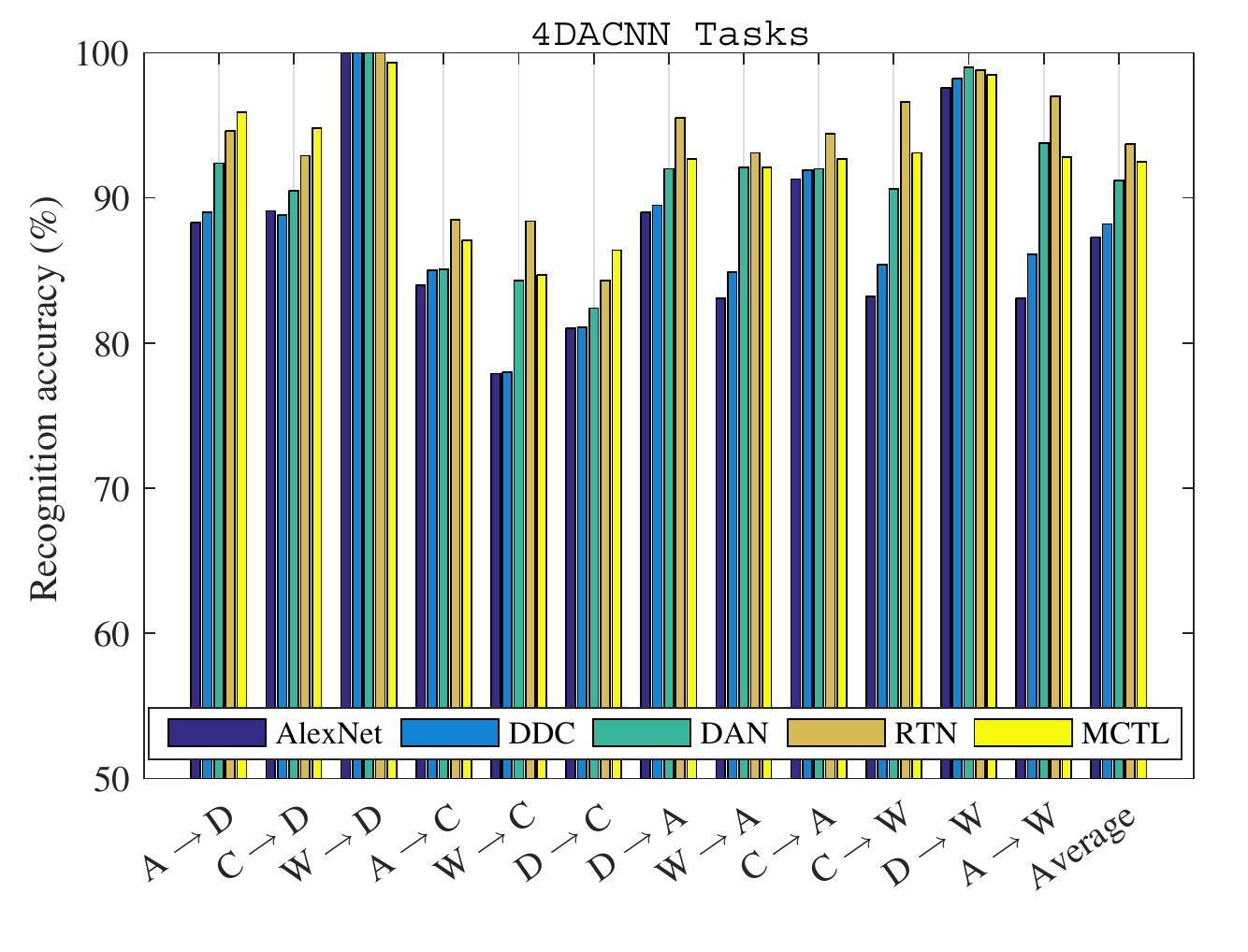}
\captionsetup{justification=centering}
   \caption{ Comparison with deep transfer learning methods}
   \label{figzhu}
\end{figure}

\begin{figure}[t]
\centering
  \includegraphics[width=0.16\linewidth,height=1.3cm]{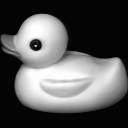}
  \includegraphics[width=0.16\linewidth,height=1.3cm]{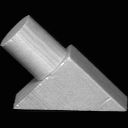}
  \includegraphics[width=0.16\linewidth,height=1.3cm]{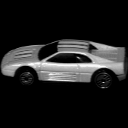}
  \includegraphics[width=0.16\linewidth,height=1.3cm]{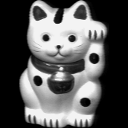}
  \includegraphics[width=0.16\linewidth,height=1.3cm]{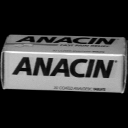}
  \includegraphics[width=0.16\linewidth,height=1.3cm]{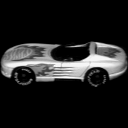}
  \includegraphics[width=0.16\linewidth,height=1.3cm]{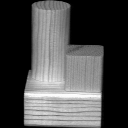}
  \includegraphics[width=0.16\linewidth,height=1.3cm]{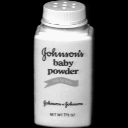}
  \includegraphics[width=0.16\linewidth,height=1.3cm]{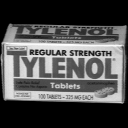}
  \includegraphics[width=0.16\linewidth,height=1.3cm]{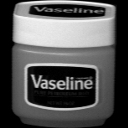}
  \includegraphics[width=0.16\linewidth,height=1.3cm]{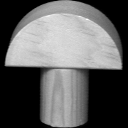}
  \includegraphics[width=0.16\linewidth,height=1.3cm]{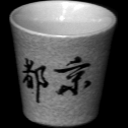}
  \includegraphics[width=0.16\linewidth,height=1.3cm]{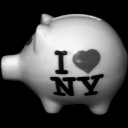}
  \includegraphics[width=0.16\linewidth,height=1.3cm]{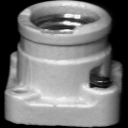}
  \includegraphics[width=0.16\linewidth,height=1.3cm]{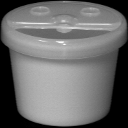}
  \includegraphics[width=0.16\linewidth,height=1.3cm]{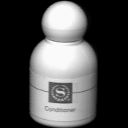}
  \includegraphics[width=0.16\linewidth,height=1.3cm]{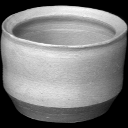}
  \includegraphics[width=0.16\linewidth,height=1.3cm]{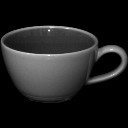}
  \includegraphics[width=0.16\linewidth,height=1.3cm]{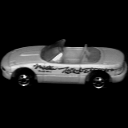}
  \includegraphics[width=0.16\linewidth,height=1.3cm]{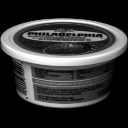}
\captionsetup{justification=centering}
   \caption{Some examples from COIL-20 dataset}
   \label{fig4}
\end{figure}

\begin{table*}\footnotesize
\captionsetup{justification=centering}
\caption{Recognition accuracy ($\%$) of different domain adaptation in 4DA Setting}
\begin{center}
\begin{tabular}{ | c | c | c | c | c | c |c | c |c |c | c |c |}
\hline

4DA Tasks&{Naive Comb}& HFA\cite{duan2012learning}&ARC-t\cite{kulis2011you}&MMDT\cite{Hoffman2014Asymmetric}&SGF\cite{Gopalan2011Domain}&GFK\cite{Gong2012Geodesic}&SA\cite{Fernando2014Unsupervised}&\tabincell{c}{LTSL\\-PCA}\cite{shao2014generalized}&\tabincell{c}{LTSL\\-LDA}\cite{shao2014generalized}&LSDT\cite{zhang2016lsdt}&$\bf{MCTL}$ \\
\hline
 $A \to D$&$55.9$&$52.7$&$50.2$&$56.7$&$46.9$&$50.9$&$55.1$&$50.4$&$\bf{59.1}$&$52.9$&$56.1$\\
\hline
 $C \to D$&$55.8$&$51.9$&$50.6$&$56.5$&$50.2$&$55.0$&$56.6$&$49.5$&$\bf{59.6}$&$56.0$&$57.3$\\
\hline
$W \to D$&$55.1$&$51.7$&$71.3$&$67.0$&$78.6$&$75.0$&$82.3$&$\bf{82.6}$&$\bf{82.6}$&$75.7$&$73.4$\\
\hline
 $A \to C$&$32.0$&$31.1$&$37.0$&$36.4$&$37.5$&$39.6$&$38.4$&$41.5$&$39.8$&$42.2$&$\bf{43.0}$\\
\hline
 $W \to C$&$30.4$&$29.4$&$31.9$&$32.2$&$32.9$&$32.8$&$34.1$&$36.7$&$\bf{38.5}$&$36.9$&$37.5$\\
\hline
 $D \to C$&$31.7$&$31.0$&$33.5$&$34.1$&$32.9$&$33.9$&$35.8$&$36.2$&$36.7$&$37.6$&$\bf{37.8}$\\
\hline
$D \to A$&$45.7$&$45.8$&$42.5$&$46.9$&$44.9$&$46.2$&$45.8$&$45.7$&$\bf{47.4}$&$46.6$&$47.0$\\
\hline
$W \to A$&$45.6$&$45.9$&$43.4$&$47.7$&$43.0$&$46.2$&$44.8$&$41.9$&$47.8$&$46.6$&$\bf{48.8}$\\
\hline
$C \to A$&$45.3$&$45.5$&$44.1$&$49.4$&$42.0$&$46.1$&$45.3$&$49.3$&$\bf{50.4}$&$47.7$&$42.8$\\
\hline
$C \to W$&$60.3$&$60.5$&$55.9$&$\bf{63.8}$&$54.2$&$57.0$&$60.7$&$50.4$&$59.5$&$57.6$&$59.6$\\
\hline
$D \to W$&$62.1$&$62.1$&$78.3$&$74.1$&$78.6$&$80.2$&$\bf{84.8}$&$81.0$&$78.3$&$83.1$&$82.1$\\
\hline
$A \to W$&$62.4$&$61.8$&$55.7$&$\bf{64.6}$&$54.2$&$56.9$&$60.3$&$52.3$&$59.5$&$57.2$&$55.7$\\
\hline
$Average$&$48.5$&$47.4$&$49.5$&$52.5$&$49.7$&$51.6$&$53.7$&$51.5$&$\bf{54.9}$&$53.3$&$54.0$\\
\hline
\end{tabular}
\end{center}
\label{tab2}
\end{table*}

\begin{table*}
\captionsetup{justification=centering}
\caption{Recognition accuracy ($\%$) of different domain adaptation of the $7^{th}$ layer in 4DACNN Setting}
\begin{center}
\begin{tabular}{ | c | c | c | c | c | c |c | c |c |}
\hline
4DA-CNN Tasks(f7)&{SourceOnly}&{Naive Comb}& SGF\cite{Gopalan2011Domain}&TCA&GFK\cite{Gong2012Geodesic}&LTSL\cite{shao2014generalized}&LSDT\cite{zhang2016lsdt}&$\bf{MCTL}$ \\
\hline
 $A \to D$&$81.3$&$94.1$&$92.0$&$82.8$&$94.3$&$94.5$&$\bf{96.0}$&$95.9$\\
\hline
 $C \to D$&$77.6$&$92.8$&$92.4$&$87.9$&$91.9$&$93.5$&$94.6$&$\bf{94.8}$\\
\hline
$W \to D$&$96.2$&$98.9$&$97.6$&$\bf{99.4}$&$98.5$&$98.8$&$99.3$&$99.3$\\
\hline
 $A \to C$&$79.3$&$83.4$&$77.4$&$81.2$&$79.1$&$85.4$&$87.0$&$\bf{87.1}$\\
\hline
 $W \to C$&$68.1$&$81.2$&$76.8$&$75.5$&$76.1$&$82.6$&$84.2$&$\bf{84.7}$\\
\hline
 $D \to C$&$74.3$&$82.7$&$78.2$&$79.6$&$77.5$&$84.8$&$86.2$&$\bf{86.4}$\\
\hline
$D \to A$&$81.8$&$90.9$&$88.0$&$90.4$&$90.1$&$91.9$&$92.5$&$\bf{92.7}$\\
\hline
$W \to A$&$73.4$&$90.6$&$86.8$&$85.6$&$85.6$&$91.0$&$91.7$&$\bf{92.1}$\\
\hline
$C \to A$&$86.5$&$90.3$&$89.3$&$92.1$&$88.4$&$90.9$&$92.5$&$\bf{92.7}$\\
\hline
$C \to W$&$67.8$&$90.6$&$87.8$&$88.1$&$86.4$&$90.8$&$\bf{93.5}$&$93.1$\\
\hline
$D \to W$&$95.1$&$98.0$&$95.7$&$96.9$&$96.5$&$97.8$&$98.3$&$\bf{98.5}$\\
\hline
$A \to W$&$71.6$&$91.1$&$88.1$&$84.4$&$88.6$&$91.5$&$\bf{92.9}$&$92.8$\\
\hline
$Average$&$79.4$&$90.4$&$87.5$&$87.0$&$87.8$&$91.1$&$92.4$&$\bf{92.5}$\\
\hline
\end{tabular}
\end{center}
\label{tab3}
\end{table*}

\begin{table*}
\captionsetup{justification=centering}
\caption{Recognition accuracy ($\%$) of different domain adaptation methods on COIL-20}
\begin{center}
\begin{tabular}{ | c | c | c | c | c | c |c | c |}
\hline
Tasks&SVM&TSL&RDALR\cite{Chang2013Robust}&DTSL\cite{Xu2015}&LTSL\cite{shao2014generalized}&LSDT\cite{zhang2016lsdt}&$\bf{MCTL}$\\
\hline
$ C1 \to C2$&$82.7$&$80.0$&$80.7$&$84.6$&$75.4$&$81.7$&$\bf{84.8}$\\
\hline
$ C2 \to C1$&$84.0$&$75.6$&$78.8$&$\bf{84.2}$&$72.2$&$81.5$&$83.7$\\
\hline
$Average$&$83.3$&$77.8$&$79.7$&$\bf{84.4}$&$73.8$&$81.6$&$84.3$\\
\hline
\end{tabular}
\end{center}
\label{tab5}
\end{table*}
\begin{figure}[t]
  \begin{minipage}[t]{0.49\linewidth} 
   \centering
    \includegraphics[width=0.3\linewidth]{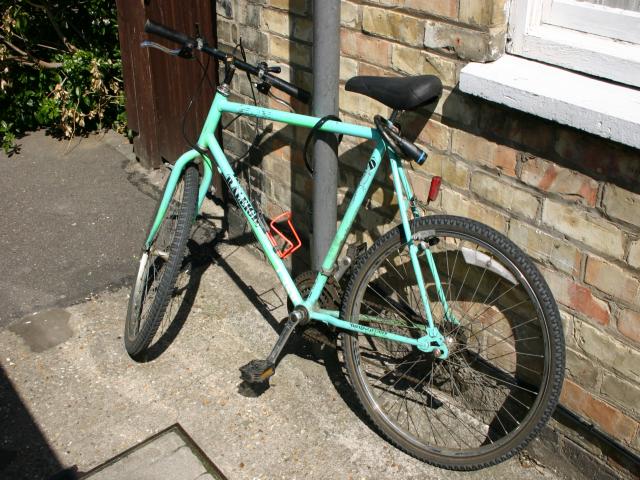}
     \includegraphics[width=0.3\linewidth]{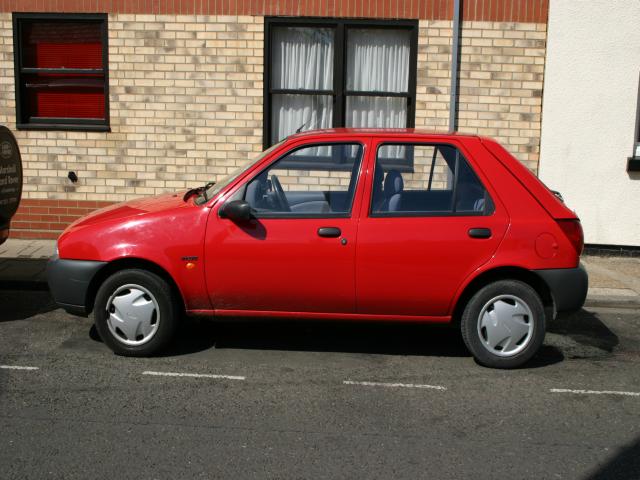}
      \includegraphics[width=0.3\linewidth]{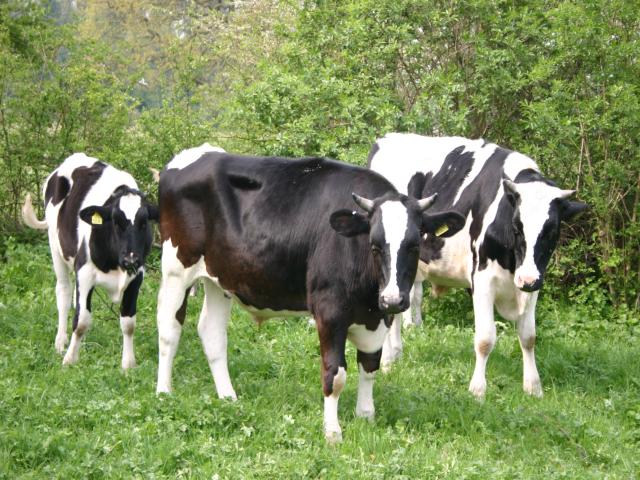}
       \includegraphics[width=0.3\linewidth]{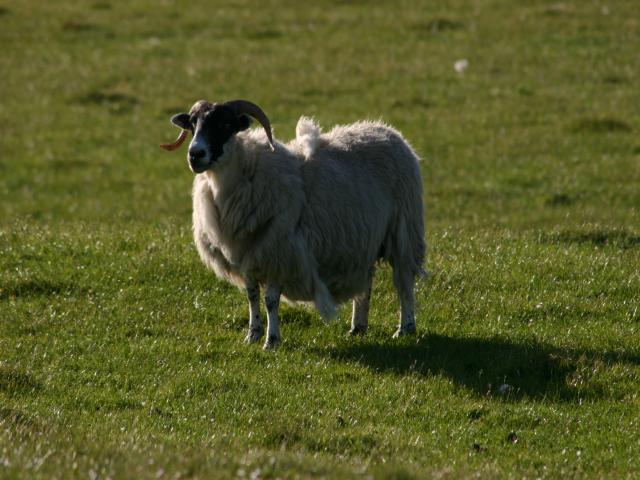}
        \includegraphics[width=0.3\linewidth]{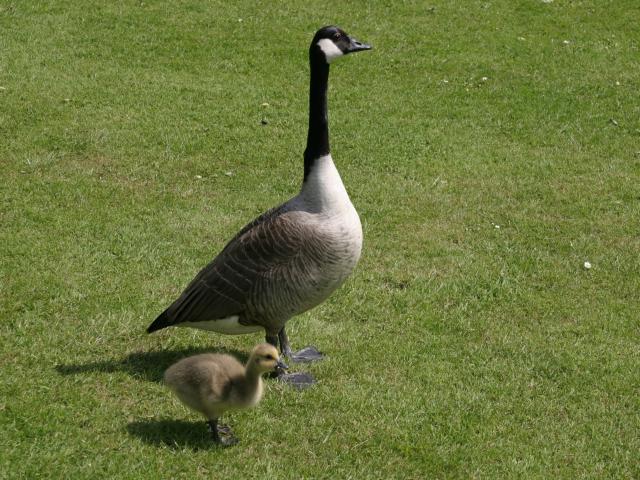}
         \includegraphics[width=0.3\linewidth]{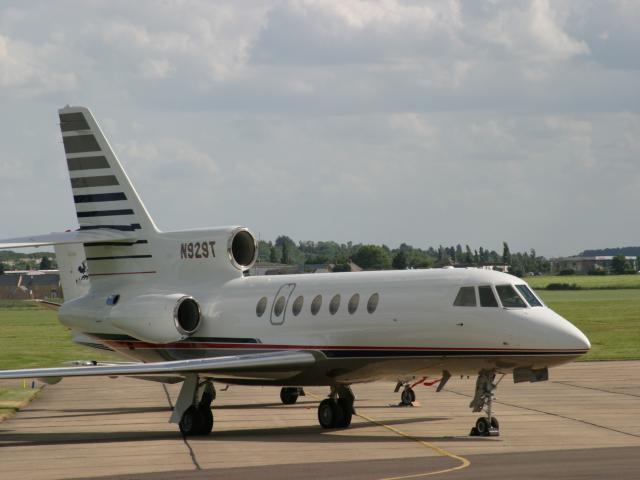}
  \caption*{ MSRC }
   \label{fig:side:a}
  \end{minipage}
  \begin{minipage}[t]{0.49\linewidth}
   \centering
    \includegraphics[width=0.3\linewidth]{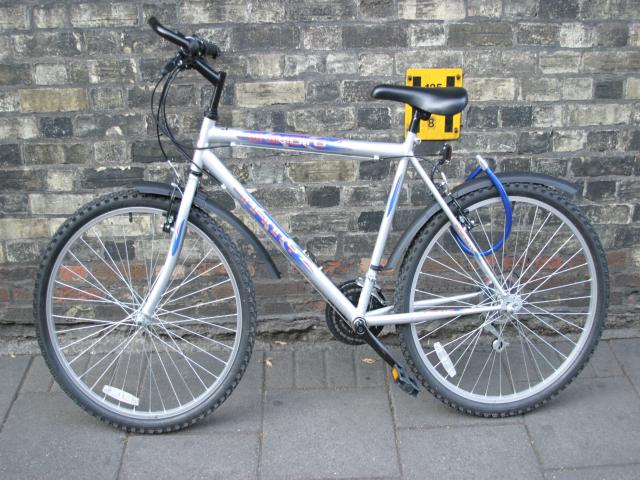}
     \includegraphics[width=0.3\linewidth]{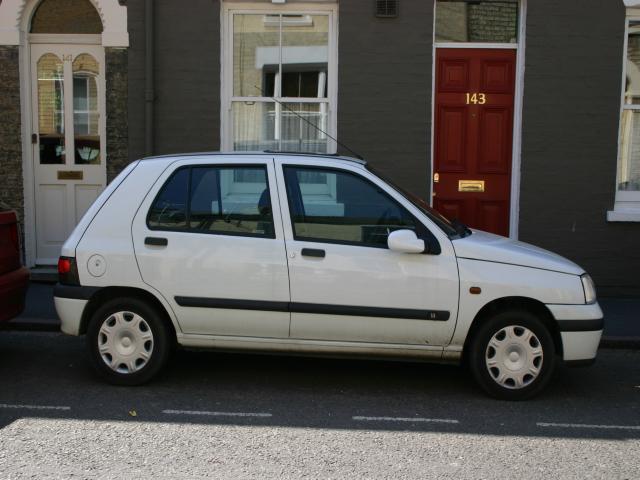}
      \includegraphics[width=0.3\linewidth]{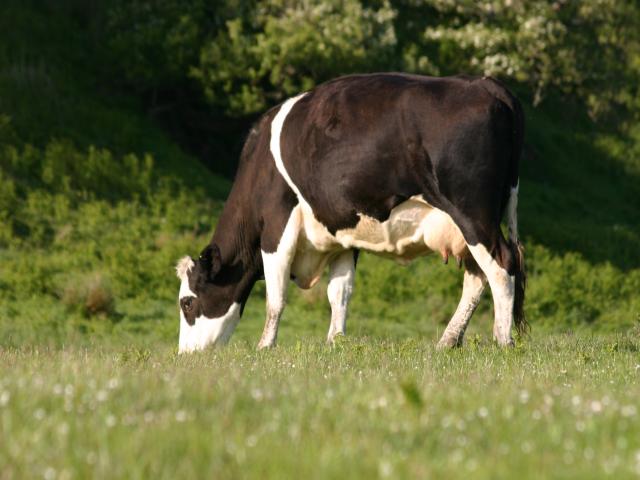}
       \includegraphics[width=0.3\linewidth]{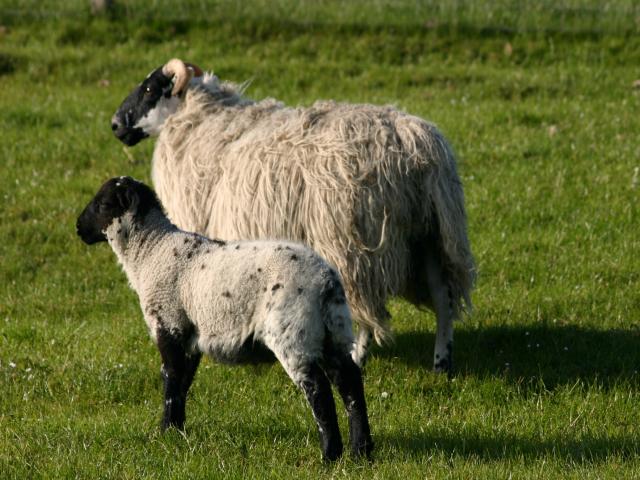}
        \includegraphics[width=0.3\linewidth]{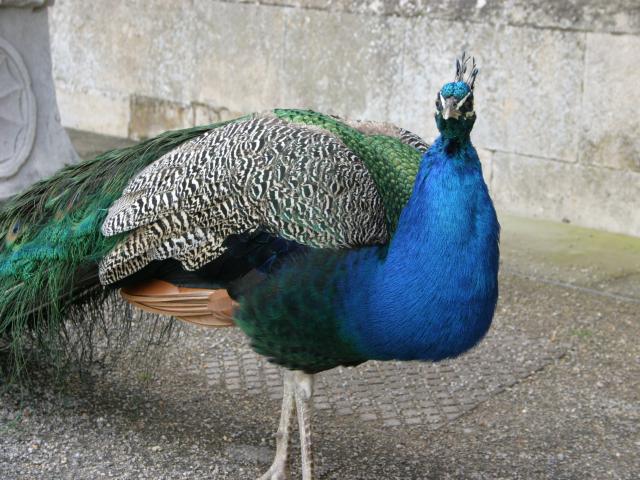}
         \includegraphics[width=0.3\linewidth]{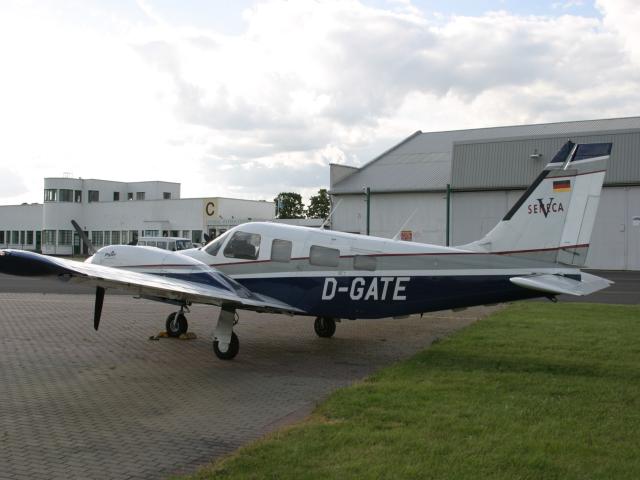}
    \caption*{VOC 2007}
    \label{fig:side:b}
  \end{minipage}
  \captionsetup{justification=centering}
   \caption{Some examples from MSRC and VOC 2007 datasets}
   \label{fig5}
\end{figure}

The recognition accuracies are reported in Table \ref{tab2}, from which we observe that the propose MCTL ranks the second ($54\%$) in average but slightly inferior to LTSL-LDA ($54.9\%$). The reason may be that the discrimination of LDA helps improve the performance, because LTSL-PCA only achieves $51.5\%$, and our MCTL also outperforms other methods. Notably, the 4DA task is a challenging benchmark, which attracts many competitive approaches for evaluation and comparison. Therefore,  excellent baselines have been achieved.

\textbf{Results on 4DA-CNN dataset (Amazon, DSLR, Webcam and Caltech 256)}\cite{Krizhevsky2012ImageNet},\cite{saenko2010eccv}:

In 4DA-CNN dataset, the CNN features are extracted by feeding the raw 4DA data (10 object classes) into the well trained convolutional neural network (AlexNet with 5 convolutional layers and 3 fully connected layers) on ImageNet \cite{Krizhevsky2012ImageNet}. The features from the $6^{th}$ and $7^{th}$ layers (i.e. DeCAF \cite{donahue2014decaf}) are explored. The feature dimensionality is 4096. In experiments, a standard configuration and protocol is used by following \cite{Gong2012Geodesic}. In this paper, the features of the $7^{th}$ layer are experimented. The recognition accuracies by using the $7^{th}$ layer outputs for 12 cross-domain tasks are shown in Table \ref{tab3}, from which we can observe that the average recognition accuracy of the proposed method shows the best performance. The superiority of generative transfer learning is demonstrated. We can see that our MCTL outperforms LTSL-LDA, this may be because there has been a better discrimination of CNN features, and discriminative learning may not significantly work.

The compared methods in Table \ref{tab3} are shallow transfer learning. It is interesting to compare with deep transfer learning methods, such as AlexNet\cite{Krizhevsky2012ImageNet}, DDC\cite{tzeng2015simultaneous}, DAN\cite{long2015learning} and RTN\cite{long2016unsupervised}. The comparison is described in Fig.\ref{figzhu}, from which we can observe that our proposed method ranks the second in average performance ($92.5\%$), which is inferior to the residual transfer network (RTN), but still better than other three deep transfer learning models. The comparison shows that the proposed MCTL, as a shallow transfer learning method, has a good competitiveness.
\begin{table*}
\captionsetup{justification=centering}
\caption{Recognition accuracy ($\%$) of different domain adaptation methods on MSRC and VOC 2007 Datasets}
\begin{center}
\begin{tabular}{ | c | c | c | c | c | c |c | c |}
\hline
Tasks&SVM&TSL&RDALR\cite{Chang2013Robust}&DTSL\cite{Xu2015}&LTSL\cite{shao2014generalized}&LSDT\cite{zhang2016lsdt}&$\bf{MCTL}$\\
\hline
$ M \to V$&$37.1$&$32.4$&$37.5$&$38.0$&$38.0$&$\bf{47.4}$&$\bf{47.4}$\\
\hline
$ V \to M$&$55.5$&$43.2$&$62.3$&$56.4$&$\bf{67.1}$&$63.9$&$64.8$\\
\hline
$Average$&$46.3$&$37.8$&$49.9$&$47.2$&$52.6$&$55.6$&$\bf{56.1}$\\
\hline
\end{tabular}
\end{center}
\label{tab6}
\end{table*}

\begin{figure}[t]
\centering
  \includegraphics[width=0.12\linewidth,height=1.2cm]{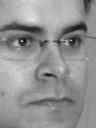}
  \includegraphics[width=0.12\linewidth,height=1.2cm]{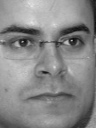}
  \includegraphics[width=0.12\linewidth,height=1.2cm]{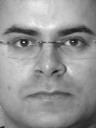}
  \includegraphics[width=0.12\linewidth,height=1.2cm]{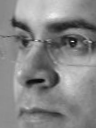}
  \includegraphics[width=0.12\linewidth,height=1.2cm]{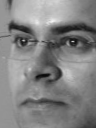}
  \includegraphics[width=0.12\linewidth,height=1.2cm]{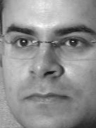}
  \includegraphics[width=0.12\linewidth,height=1.2cm]{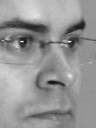}
  \includegraphics[width=0.12\linewidth,height=1.2cm]{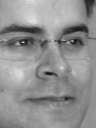}
  \includegraphics[width=0.12\linewidth,height=1.2cm]{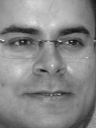}
  \includegraphics[width=0.12\linewidth,height=1.2cm]{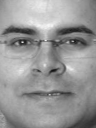}
  \includegraphics[width=0.12\linewidth,height=1.2cm]{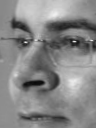}
  \includegraphics[width=0.12\linewidth,height=1.2cm]{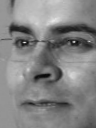}
  \includegraphics[width=0.12\linewidth,height=1.2cm]{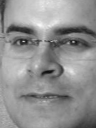}
  \includegraphics[width=0.12\linewidth,height=1.2cm]{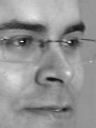}
\captionsetup{justification=centering}
   \caption{Facial images of one person from CMU Multi-PIE}
   \label{fig6}
\end{figure}
\begin{figure}[t]
\centering
  \includegraphics[width=0.8\linewidth]{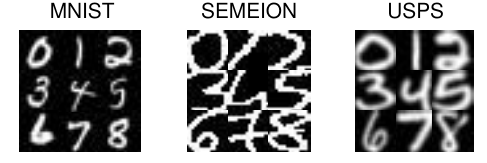}
\captionsetup{justification=centering}
   \caption{ Some images from handwritten digits datasets}
   \label{fig7}
\end{figure}

\textbf{Results on COIL-20 dataset\footnote{\url{http://www.cs.columbia.edu/CAVE/software/softlib/coil-20.php}}: Columbia Object Image Library\cite{Rate2011Columbia}}:

The COIL-20 dataset contains 20 objects with 1440 gray scale images (72 multi-pose images per object). The image size is $128\times 128$ of 256 gray levels. In experiments, by following the experimental protocol in\cite{Xu2015}, the size of each image is cropped into $32\times 32$ and the dataset is divided into two subsets C1 and C2, with each 2 quadrants are included. Specifically, the C1 set contains the directions of [$0^\circ$, $85^\circ$] and [$180^\circ$, $265^\circ$], from quadrants 1 and 3. The C2 set contains the directions of [$90^\circ$, $175^\circ$] and [$270^\circ$, $355^\circ$], from quadrants 2 and 4. The two subsets are distribution different but relevant in semantic, and therefore come to a DA problem. By taking C1 and C2 as source and target domain alternatively, the cross-domain recognition rates of different methods are shown in Table \ref{tab5}, from which we see that the proposed MCTL ($84.3\%$) is a little inferior to DTSL ($84.4\%$), but shows a superior performance over other related methods, especially the recent LSDT method ($81.6\%$).

\textbf{Results on MSRC\footnote{\url{http://research.microsoft.com/en-us/projects/objectclassrecognition}} and VOC 2007\footnote{\url{http://pascallin.ecs.soton.ac.uk/challenges/VOC/voc2007}} datasets:\cite{Xu2015}}:

The MSRC dataset contains 4323 images with 18 classes and the VOC 2007 dataset contains 5011 images with 20 concepts. The two datasets share 6 semantic classes: airplane, bicycle, bird, car, cow and sheep. We follow\cite{long2014transfer} to construct a cross-domain image dataset MSRC vs. VOC ($M\to V$) by selecting 1269 images from MSRC as the source domain, and 1530 images from VOC 2007 as the target domain. Then we switch the two datasets: VOC vs. MSRC ($V\to M$). All images are uniformly rescaled to 256 pixels, and 128-dimensional dense SIFT (DSIFT) features using the VLFeat open source package are extracted. Then $K$-means clustering is used to obtain a 240-dimensional codebook. In this way, the source and target domain data are constructed to share the same label set. The experimental results of different domain adaptation methods are shown in Table \ref{tab6}, from which we observe that the performance of our method is $0.5\%$ higher than state-of-the-art LSDT method and $3.5\%$ higher than LTSL method in average cross-domain recognition performance.

\subsection{Cross-poses Face Recognition}
It is known that 3D pose change in faces is a nonlinear transfer problem, general recognition models are very sensitive to pose change. Therefore, it is challenging to handle the pose based face recognition issue.
In this section, the popular CMU Multi-PIE face dataset\footnote{\url{http://www.cs.cmu.edu/afs/cs/project/PIE/MultiPie/Multi-Pie/Home.html}} with 337 subjects is used. Each subject contains 4 different sessions with 15 poses, 20 illuminations, and 6 expressions. The facial images in Session 1 and Session 2 of one person are shown in Fig. \ref{fig6}. In our experiment, we select the first 60 subjects from Session 1 and Session 2. As a result, a smaller session 1 (S1) with 7 images of different poses per class under neutral expression and a smaller session 2 (S2) that is similar to S1 but under smile expression are constructed as domain data. Notably, the raw image pixels are used as features. Specifically, the experimental configurations are set as follows.

\textbf{S1}: One frontal face ($0^\circ$) per subject is used as source data, one $60^\circ$ posed face is used as the target training data, and the remaining 5 facial images are used as the target test data.

\textbf{S2}: The experimental configuration is the same as S1.

\textbf{S1+S2}: The two frontal faces ($0^\circ$) and the two $60^\circ$ posed faces under neutral and smile expression are used as source data and target training data in the two sessions, respectively. The remaining 10 facial images are used as target test data.

\textbf{S1 $\to$ S2}: S1 is used as source data, the frontal and $60^\circ$ posed faces in S2 are used as the target training data, and the remaining data in S2 are used as test data.

With above settings, the recognition accuracies of different methods have been shown in Table \ref{tab7}. It is clear that the proposed method performs significantly better, which is $5\%$ over other DA methods in handling such pose variation based nonlinear transfer problem. This also demonstrates that the proposed intermediate domain generation based transfer learning can better interpret local generative discrepancy metric (LGDM) and improve the nonlinear local transfer problem. The manifold criterion is then validated.

\begin{table*}
\captionsetup{justification=centering}
\caption{Recognition accuracy ($\%$) of different domain adaptation methods on face recognition across poses}
\begin{center}
\begin{tabular}{ | c | c | c | c | c | c |c | c |c |}
\hline
Tasks&{Naive Comb}&{A-SVM}& SGF\cite{Gopalan2011Domain} &GFK\cite{Gong2012Geodesic}&SA\cite{Fernando2014Unsupervised}&{LTSL\cite{shao2014generalized}}&LSDT\cite{zhang2016lsdt}&$\bf{MCTL}$ \\
\hline
$S1$ ($0^\circ \to 60^\circ$) &$61.0$&$57.0$&$53.7$&$61.0$&$51.3$&$56.0$&$59.7$&$\bf{65.3}$\\
\hline
$S2$ ($0^\circ\to60^\circ$) &$62.7$&$62.7$&$55.0$&$58.7$&$62.7$&$62.7$&$63.3$&$\bf{70.0}$\\
\hline
$S1+S2$ ($0^\circ\to60^\circ$) &$60.2$&$60.1$&$53.8$&$56.3$&$61.7$&$60.2$&$61.7$&$\bf{68.3}$\\
\hline
$S1\to S2$&$93.6$&$94.3$&$92.5$&$96.7$&$98.3$&$97.2$&$95.8$&$\bf{98.7}$\\
\hline
$Average$&$69.4$&$68.5$&$63.8$&$67.0$&$68.5$&$70.3$&$70.1$&$\bf{75.6}$\\
\hline
\end{tabular}
\end{center}
\label{tab7}
\end{table*}

\begin{table*}
\captionsetup{justification=centering}
\caption{Recognition accuracy ($\%$) of different domain adaptation on handwritten digits recognition}
\begin{center}
\begin{tabular}{ | c | c | c | c | c | c |c | c |c |}
\hline
 Tasks&{Naive Comb}&{A-SVM}& SGF\cite{Gopalan2011Domain}&GFK\cite{Gong2012Geodesic}&SA\cite{Fernando2014Unsupervised}&LTSL\cite{shao2014generalized}&LSDT\cite{zhang2016lsdt}&$\bf{MCTL}$ \\
\hline
 $M \to U$&$78.8$&$78.3$&$79.2$&$82.6$&$78.8$&$83.2$&$79.3$&$\bf{87.8}$\\
\hline
 $S \to U$&$83.6$&$76.8$&$77.5$&$82.7$&$82.5$&$83.6$&$84.7$&$\bf{84.8}$\\
\hline
$M \to S$&$51.9$&$70.5$&$51.6$&$70.5$&$\bf{74.4}$&$72.8$&$69.1$&$74.0$\\
\hline
 $U \to S$&$65.3$&$74.5$&$70.9$&$76.7$&$74.6$&$65.3$&$67.4$&$\bf{83.0}$\\
\hline
 $U \to M$&$71.7$&$73.2$&$71.1$&$74.9$&$72.9$&$71.7$&$70.5$&$\bf{81.2}$\\
\hline
 $S \to M$&$67.6$&$69.3$&$66.9$&$74.5$&$72.9$&$67.6$&$70.0$&$\bf{74.0}$\\
 \hline
 $Average$&$69.8$&$73.8$&$69.5$&$77.0$&$76.0$&$74.0$&$73.5$&$\bf{80.8}$\\
\hline
\end{tabular}
\end{center}
\label{tab8}
\end{table*}

\begin{table}[t]
\caption{Average performance of all transfer tasks}
\begin{center}
\begin{tabular}{ | c | c | c | c |}
\hline
All Transfer Tasks&LTSL\cite{shao2014generalized}& LSDT\cite{zhang2016lsdt}&MCTL \\
\hline
 $Average~($\%$)$&$69.45$&$71.08$&$73.88$\\
\hline
\end{tabular}
\end{center}
\label{tab15}
\end{table}

\subsection{Cross-domain Handwritten Digits Recognition}
Three handwritten digits datasets including MNIST (M)\footnote{\url{http://yann.lecun.com/exdb/mnist/}}, USPS (U)\footnote{\url{http://www-i6.informatik.rwth-aachen.de/~keysers/usps.html}} and SEMEION (S)\footnote{\url{http://archive.ics.uci.edu/ml/datasets/Semeion+Handwritten+Digit}} with 10 classes from digit $0\sim9$ are used for evaluating the proposed MCTL. The MNIST dataset consists of 70,000 instances of $28\times28$, the USPS dataset consists of 9298 examples of $16\times16$, and the SEMEION dataset consists of 2593 images of $16\times16$. The MNIST dataset is cropped into $16\times16$. Several images from three datasets are shown in Fig. \ref{fig7}.
Each dataset is used as source and target domain alternatively, and 6 cross-domain tasks are explored. Also, 100 samples per class from source domain and 10 samples per class from target domain are randomly selected for training. 5 random splits are used, and the average classification accuracies are reported in Table \ref{tab8}.
From the results, we observe that our MCTL outperforms other state-of-the-art methods with $3\%$, and the significant superiority is therefore proved.

From the whole experiments on 4DA, 4DA-CNN, COIL-20, MSRC and VOC2007, Multi-PIE, and Handwritten digits, we can see that the proposed MCTL shows competitive performance. Although our MCTL shows very slight improvement on several tasks by comparing to state-of-the-art method, the comprehensive superiority of MCTL in all datasets is clearly demonstrated in Table \ref{tab15}, which shows the mean value of all the cross-domain tasks in the datasets. From the results, we can observe that our MCTL outperforms state-of-the-art LTSL and LSDT about 2.8\% in average performance on all the transfer tasks explored in this paper.

\section{Discussion}
\subsection{Analysis of MCTL-S}

When the condition ${\textbf{X}_{GT}}={\textbf{X}_T}$ is strictly satisfied, i.e. perfect domain generation, our model is degenerated into the MCTL-S model, which can be simply formulated as problem (\ref{fun17}). The MC-S loss is more similar to a generic manifold regularization, which is built in an ideal condition focusing on the locality structure. Under this case, domain generation relies more on local manifold, regardless of the global property. Therefore, the performance of the MCTL-S with ideal and perfect condition will degrade when global shift of domain data is encountered. The GGDM loss that measures the global structure can be an effective relaxation.

The experimental comparisons on 4DACNN dataset between MCTL and MCTL-S are presented in Table \ref{tab12} and the comparisons on COIL-20 dataset are shown in Table \ref{tab13}. From the results, we observe that the proposed MCTL and the harsh MCTL-S performs similar performance. This demonstrates that domain generation is a feasible way for unsupervised domain transfer learning. It is also encouraging for us to use deep generative method (e.g. GAN) for transfer learning in the future. The potential problem of GAN is that the similar high-level semantic information across domain may be generated, but the distribution may still be inconsistent.

\subsection{Parameter Setting and Ablation Analysis}

In our method, the trade-off coefficients $\tau$ and ${\lambda _1}$ are fixed as 1 in experiments. Dimensions of common subspace is set as $d=n$.
The Gaussian kernel function ${k}(\mathbf{x}_i,\mathbf{x}_j)=$exp$(-\parallel\mathbf{x}_i-\mathbf{x}_j\parallel^2/2\sigma^2)$ is used, where $\sigma$ can be tuned for different tasks, e.g. $\sigma=1.2$ for 4DA-CNN and $\sigma=0.8$ for COIL-20. But the linear kernel function is adopted for discussion as it can effectively avoid the influence of kernel parameter. The least square classifier\cite{kanamori2009least} is used in DA experiments except that in COIL-20 experiment, the SVM classifier is used because of its good performance.

\begin{table}[h]
\caption{Recognition accuracy ($\%$) in 4DACNN dataset}
\begin{center}
\begin{tabular}{ | c | c | c |}
\hline
4DA-CNN Tasks&MCTL& MCTL-S \\
\hline
 $A \to D$&$95.67$&$95.71$\\
\hline
 $C \to D$&$94.69$&$94.72$\\
\hline
$W \to D$&$99.25$&$99.29$\\
\hline
 $A \to C$&$87.11$&$87.05$\\
\hline
 $W \to C$&$84.73$&$84.74$\\
\hline
 $D \to C$&$86.37$&$86.34$\\
\hline
$D \to A$&$92.66$&$92.65$\\
\hline
$W \to A$&$92.06$&$92.07$\\
\hline
$C \to A$&$92.68$&$92.06$\\
\hline
$C \to W$&$93.08$&$93.04$\\
\hline
$D \to W$&$98.49$&$98.51$\\
\hline
$A \to W$&$92.79$&$92.83$\\
\hline
$Average$&$92.47$&$92.47$\\
\hline
\end{tabular}
\end{center}
\label{tab12}
\end{table}

\begin{table}[h]
\caption{Recognition accuracy ($\%$) in COIL-20}
\begin{center}
\begin{tabular}{ | c | c | c |}
\hline
COIL-20&MCTL& MCTL-S \\
\hline
 $ C1 \to C2$&$84.83$&$85.00$\\
\hline
 $ C2 \to C1$&$83.67$&$83.67$\\
\hline
Average&$84.25$&$84.34$\\
\hline
\end{tabular}
\end{center}
\label{tab13}
\end{table}

In MCTL model, three items such as MMD loss based GGDM term, MC loss based LGDM term and LRC regularization term are included. For better interpreting the effect of each term, the ablation analysis by removing one of them is discussed. Therefore, some extra experiments on the COIL-20 object recognition task (i.e. $C1 \to C2$), Handwritten Digits recognition task (i.e. $M \to U$) and  MSRC-VOC 2007 image recognition task (i.e. $V \to M$) are studied for ablation analysis. The experimental results are shown in Table \ref{tab14}. We can observe that the LGDM loss plays more important role than GGDM loss, with $2.4\%$ improvement in average. This is reasonable because in many real cross-domain tasks, global transfer may result in negative transfer, due to the local bias problem of domain discrepancy. This further demonstrates the superiority and validity of the proposed MCTL because local discrepancy metric is deserved for transfer learning.

\subsection{Model Dimensionality and Parameter Analysis}

\begin{figure}
\begin{center}
 \includegraphics[width=1\linewidth]{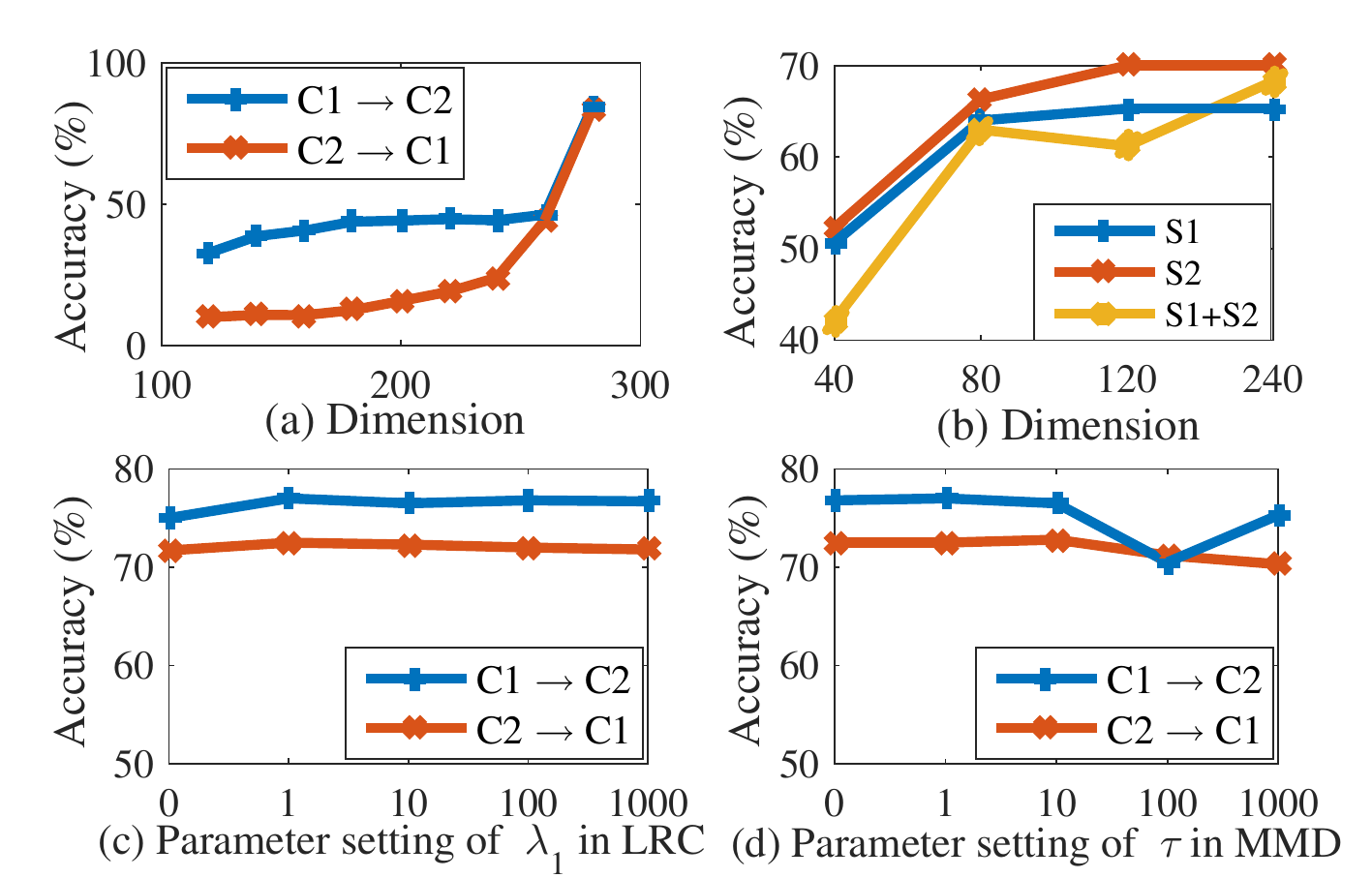}
\end{center}
\captionsetup{justification=centering}
   \caption{Dimensionality and parameter sensitivity analysis}
   \label{fig9}
\end{figure}

\textbf{Dimensionality Analysis}. In MCTL model, a latent common subspace $\bm{\mathcal {P}}$ is learned. Therefore,  the performance variation with different subspace dimensions is studied on the COIL-20 ($C1 \to C2$ and $C1 \to C2$) and CMU Multi-PIE face datasets including $S1$, $S2$, and $S1+S2$ tasks. The performance curve with increasing number of the dimensionality $d$ is shown in Fig. \ref{fig9} (a) and (b). Generally, the recognition performance can be improved with increasing number of dimension.

\textbf{Parameter Sensitivity Analysis}. In MCTL model, there are two trade-off parameters $\tau$ and ${\lambda _1}$ involved in parameter tuning. To have an insight of their sensitivity to model, the parameter sensitivity analysis is studied on COIL-20 ($C1\to C2$ and $C2\to C1$) task by tuning the parameters from $\{0,1,10,100,1000\}$, respectively. Fig. \ref{fig9} (c) shows the parameter analysis of $\lambda _1$ by fixing ${\tau}=1$. Fig. \ref{fig9} (d) shows the parameter analysis of $\tau$ by fixing ${\lambda _1}=1$. For tuning the two parameters simultaneously, we have also provided the 3D surface on COIL-20 dataset in Fig.\ref{fig14} (a) ($C1\to C2$) and Fig.\ref{fig14} (b) ($C2\to C1$). We can see that the model is robust to the model parameters, without serious fluctuation.
\begin{small}
\begin{table}
\caption{Results of ablation analysis}
\begin{center}
\begin{tabular}{ | c | c | c | c | c |}
\hline
Tasks& MCTL &no LGDM &no LRC&no GGDM\\
\hline
$ C1 \to C2$&$\bf{77.0}$&$73.0$&$76.7$&$76.8$\\
\hline
$ M \to U$&$71.0$&$70.0$&$67.0$&$\bf{73.0}$\\
\hline
$ V \to M$&$70.2$&$70.1$&$70.1$&$\bf{70.3}$\\
\hline
 $Average$&$72.7$&$71.0$&$71.2$&$\bf{73.4}$\\
\hline
\end{tabular}
\end{center}
\label{tab14}
\end{table}
\end{small}

\begin{table*}\small
\captionsetup{justification=centering}
\caption{Computational time analysis and recognition accuracy ($\%$)}
\begin{center}
\begin{tabular}{ | c | c | c | c | c | c |}
\hline
 Tasks& SGF\cite{Gopalan2011Domain}&GFK\cite{Gong2012Geodesic}&SA\cite{Fernando2014Unsupervised}&LTSL\cite{shao2014generalized}&$\bf{MCTL}$ \\
\hline
$S1 \to S2$&$10.9s$ ($92.5\%$)&$1.5s$ ($96.7\%$)&$4.18s$ ($98.3\%$)&$7.21s$ ($97.2\%$)&$7.62s$ ($97.3\%$)\\
\hline
$M \to U$&$75s$ ($79.2\%$)&$12.2s$ ($82.6\%$)&$30.5s$ ($78.8\%$)&$62.1s$ ($83.2\%$)&$98.8s$ ($87.8\%$)\\
\hline
\end{tabular}
\end{center}
\label{tab11}
\end{table*}

\subsection{Computational Complexity and Time Analysis}

In this section, the computational complexity of the Algorithm 1 is presented. The algorithm includes three basic steps: update $\bm{\mathcal {Z}}$, update $\bm{\mathcal{J}}$, and update $\mathbf\Phi$. The computation of $\mathbf\Phi$ involves eigen-decomposition and matrix multiplication, and the complexity is $O(n^3)$. The computation of updating $\bm{\mathcal{J}}$ and $\bm{\mathcal{Z}}$ is $O(n^2)$. Suppose that the number of iterations is $T$, then the total computational complexity of MCTL can be expressed as $O(T n^3)+ O(T n^2)$. It is noteworthy that the complexity of Gram matrix computation is not included, because it can be computed in advance without computing in Algorithm 1.

Further, Table \ref{tab11} shows the computational time comparisons on CMU Multi-PIE data ($S1 \to S2$) and handwritten digits data ($M \to U$). From Table \ref{tab11}, we observe that the proposed MCTL has also a low computational time. We should claim that the proposed method is better used together with deep models for large-scale data, due to the stronger feature representation capability of deep methods with large-scale data. Notably, all algorithms in experiments are implemented in computer of Intel i5-4460 CPU, 3.20GHz, and 16GB RAM.

\subsection{Model Visualization and Convergence}

\begin{figure}[t]
\begin{center}
 \includegraphics[width=0.9\linewidth]{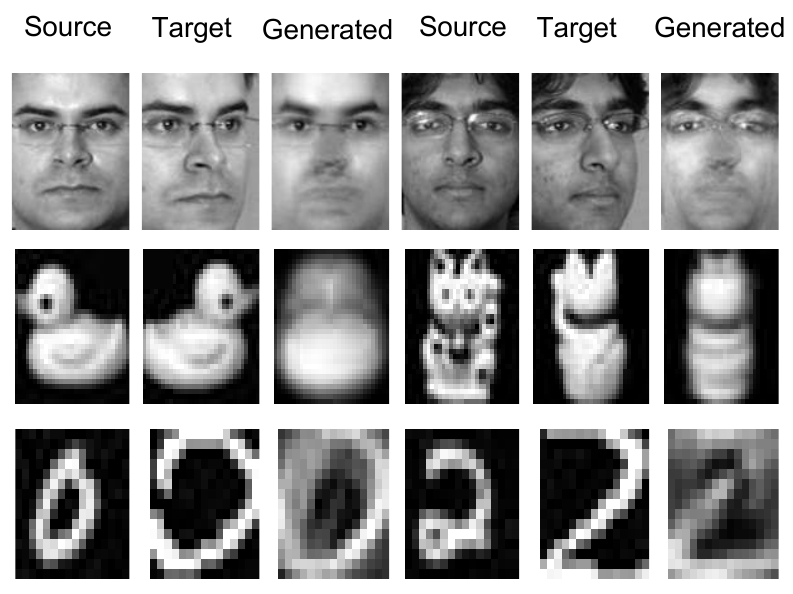}
\end{center}
\captionsetup{justification=centering}
   \caption{Visualization of MCTL alignment}
   \label{fig11}
\end{figure}

\begin{figure}[t]
\begin{center}
\includegraphics[width=1\linewidth]{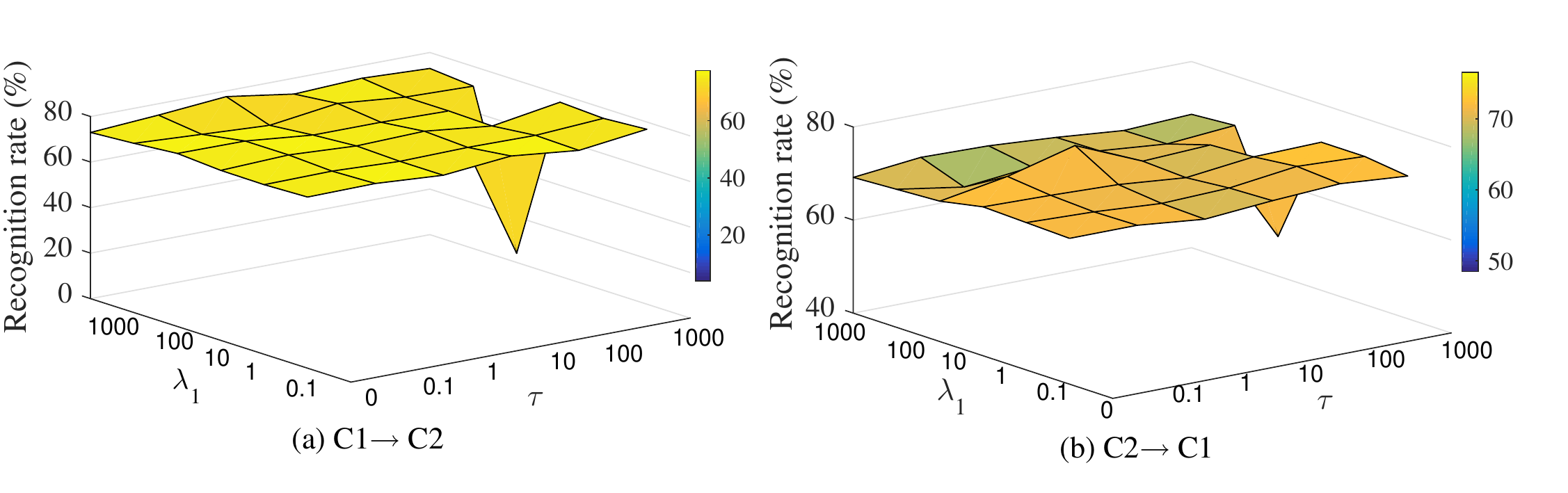}
\end{center}
\captionsetup{justification=centering}
   \caption{Parameter sensitivity analysis}
   \label{fig14}
\end{figure}

In this section, the visualization and convergence will be discussed. Pose alignment is a difficult task. Therefore, for better insight of the MCTL model, the feature visualization is explored. We have shown the visualization of CMU PIE. The first row in Fig. \ref{fig11} illustrates the pose transfer process under Session 1 via MCTL, from which we observe that the generated intermediate domain data by source data inherits similar distribution property of target data.

Further, COIL-20 and handwritten digits datasets are also exploited. The second row of Fig. \ref{fig11} shows the pose transfer process, and the generative data shows a compromise of source and target data in visual disparity. Similarly, the visualization of the generated handwritten digits (intermediate domain) by taking MNIST as the source domain and SEMEION as target domain is shown in the third row of Fig. \ref{fig11}. The effect of domain generation is clearly shown.

\begin{figure}[t]
\begin{center}
 \includegraphics[width=1\linewidth]{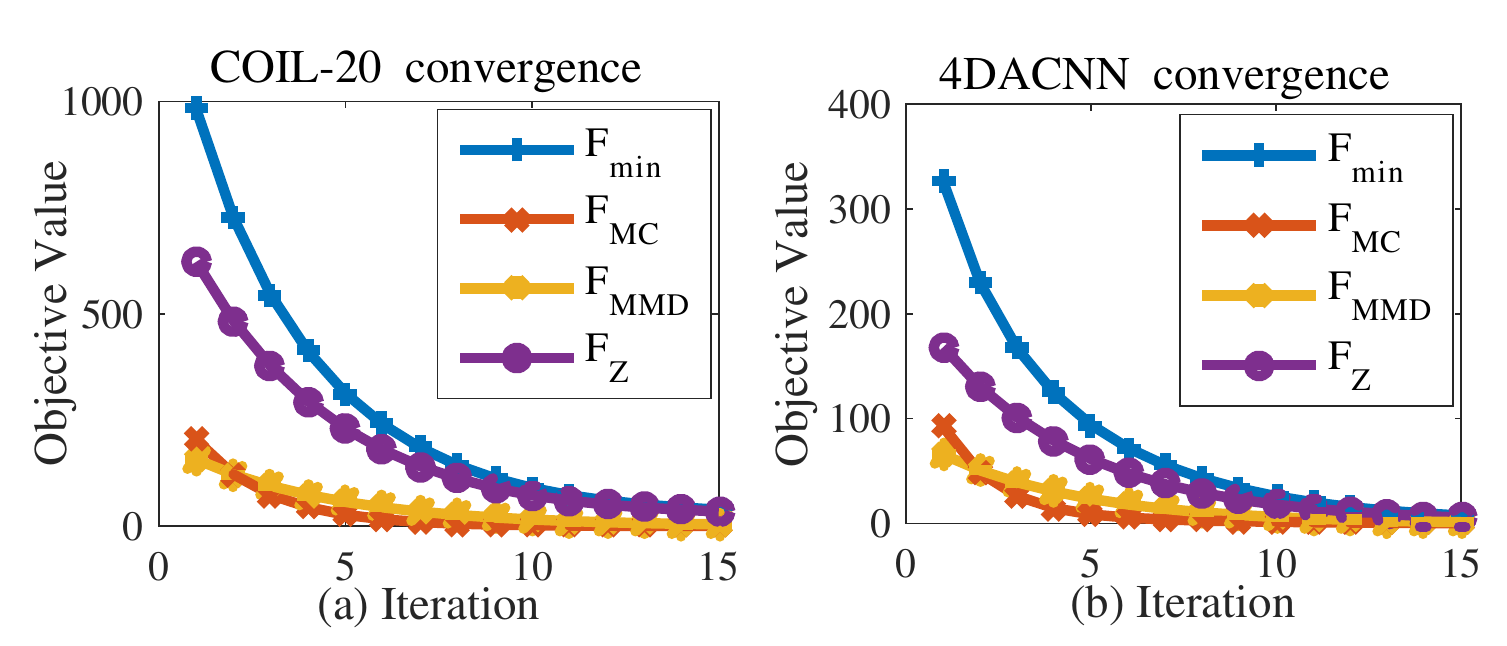}
\end{center}
\captionsetup{justification=centering}
   \caption{Convergence of MCTL algorithm}
   \label{fig13}
\end{figure}

Additionally, the convergence of our MCTL method is explored by observing the variation of the objective function. In the experiments, the number of iterations is set to be 15, and the variation of the objective function (i.e. $F_{min}$) is described in Fig. \ref{fig13}. It is clear that the objective function decreases to a constant value after several iterations, by running the algorithm, on COIL-20 ($C1\to C2$) and 4DACNN ($A \to D$), respectively. Also, the convergence of each term in the MCTL, such as $F_{MC}$ (i.e. MC based LGDM loss), $F_{MMD}$ (i.e. GGDM loss), and $F_{\bm{\mathcal {Z}}}$ (i.e. LRC regularization) are also presented in Fig. \ref{fig13}. We can observe the fast convergence of MCTL after several iterations. Notably, the optimization solver in this paper may not be optimal selection, and the performance may be further fine-tuned with better solvers.

\section{Conclusion}

In this paper, we propose a new transfer learning perspective with intermediate domain generation. Specifically, a Manifold Criterion Guided Transfer Learning (MCTL) method is introduced. In previous work, MMD is commonly used for global domain discrepancy minimization and achieves good performance in domain adaptation. However, an open problem, that MMD neglects the locality geometric structure of domain data, is preserved. In order to overcome the bottleneck, motivated by manifold criterion, MCTL is proposed, which aims at generating a new intermediate domain sharing similar distribution with the true target domain. The manifold criterion (MC) implies that the domain adaptation is achieved if MC is satisfied (i.e. minimal domain discrepancy). The rationale behind MC is that if the locality structure is preserved between the generated intermediate domain and the true target domain, then the $i.i.d.$ condition is achieved. Finally, with a MC based LGDM loss, GGDM loss and LRC regularization jointly constructed, MCTL is established. Extensive experiments on benchmark DA datasets demonstrate the superiority of the proposed method over several state-of-the-art DA methods.


%

%

\ifCLASSOPTIONcompsoc
  \section*{Acknowledgments}
\else
  \section*{Acknowledgment}
\fi

The authors would like to thank the editor and the anonymous reviewers for their valuable comments and suggestions.

\ifCLASSOPTIONcaptionsoff
  \newpage
\fi



%
{\normalsize
\bibliographystyle{./IEEEtran}
\bibliography{egbib}

\begin{thebibliography}{10}
\providecommand{\url}[1]{#1}
\csname url@samestyle\endcsname
\providecommand{\newblock}{\relax}
\providecommand{\bibinfo}[2]{#2}
\providecommand{\BIBentrySTDinterwordspacing}{\spaceskip=0pt\relax}
\providecommand{\BIBentryALTinterwordstretchfactor}{4}
\providecommand{\BIBentryALTinterwordspacing}{\spaceskip=\fontdimen2\font plus
\BIBentryALTinterwordstretchfactor\fontdimen3\font minus
  \fontdimen4\font\relax}
\providecommand{\BIBforeignlanguage}[2]{{%
\expandafter\ifx\csname l@#1\endcsname\relax
\typeout{** WARNING: IEEEtran.bst: No hyphenation pattern has been}%
\typeout{** loaded for the language `#1'. Using the pattern for}%
\typeout{** the default language instead.}%
\else
\language=\csname l@#1\endcsname
\fi
#2}}
\providecommand{\BIBdecl}{\relax}
\BIBdecl

\bibitem{Nguyen2015DASH}
H.~V. Nguyen, H.~T. Ho, V.~M. Patel, and R.~Chellappa, ``Dash-n: Joint
  hierarchical domain adaptation and feature learning,'' \emph{IEEE Trans.
  Image Process}, vol.~24, no.~12, pp. 5479--5491, 2015.

\bibitem{csurka2017domain}
G.~Csurka, ``Domain adaptation for visual applications: A comprehensive
  survey,'' \emph{arXiv}, 2017.

\bibitem{hoffman2012discovering}
J.~Hoffman, B.~Kulis, T.~Darrell, and K.~Saenko, ``Discovering latent domains
  for multisource domain adaptation,'' in \emph{ECCV}.\hskip 1em plus 0.5em
  minus 0.4em\relax Springer, 2012, pp. 702--715.

\bibitem{pan2010survey}
S.~J. Pan and Q.~Yang, ``A survey on transfer learning,'' \emph{IEEE Trans.
  Knowle. Data Engineering}, vol.~22, no.~10, pp. 1345--1359, 2010.

\bibitem{Duan2012Domain}
L.~Duan, I.~W. Tsang, and D.~Xu, ``Domain transfer multiple kernel learning,''
  \emph{IEEE Trans. PAMI}, vol.~34, no.~3, pp. 465--479, 2012.

\bibitem{li2017domain}
W.~Li, Z.~Xu, D.~Xu, D.~Dai, and L.~Van~Gool, ``Domain generalization and
  adaptation using low rank exemplar svms,'' \emph{IEEE Trans. PAMI}, 2017.

\bibitem{gopalan2014unsupervised}
R.~Gopalan, R.~Li, and R.~Chellappa, ``Unsupervised adaptation across domain
  shifts by generating intermediate data representations,'' \emph{IEEE Trans.
  PAMI}, vol.~36, no.~11, pp. 2288--2302, 2014.

\bibitem{Duan2010Visual}
L.~Duan, D.~Xu, I.~W. Tsang, and J.~Luo, ``Visual event recognition in videos
  by learning from web data,'' in \emph{CVPR}, 2010, pp. 1959--1966.

\bibitem{kulis2011you}
B.~Kulis, K.~Saenko, and T.~Darrell, ``What you saw is not what you get: Domain
  adaptation using asymmetric kernel transforms,'' in \emph{CVPR}, 2011, pp.
  1785--1792.

\bibitem{saenko2010adapting}
K.~Saenko, B.~Kulis, M.~Fritz, and T.~Darrell, ``Adapting visual category
  models to new domains,'' 2010.

\bibitem{pan2011domain}
S.~J. Pan, I.~W. Tsang, J.~T. Kwok, and Q.~Yang, ``Domain adaptation via
  transfer component analysis,'' \emph{IEEE Trans. Neural Networks}, vol.~22,
  no.~2, pp. 199--210, 2011.

\bibitem{Duan2009Domain}
L.~Duan, I.~W. Tsang, D.~Xu, and S.~J. Maybank, ``Domain transfer svm for video
  concept detection,'' in \emph{CVPR}, 2009, pp. 1375--1381.

\bibitem{baktashmotlagh2013unsupervised}
M.~Baktashmotlagh, M.~T. Harandi, B.~C. Lovell, and M.~Salzmann, ``Unsupervised
  domain adaptation by domain invariant projection,'' in \emph{ICCV}, 2013, pp.
  769--776.

\bibitem{ben2007analysis}
S.~Ben-David, J.~Blitzer, K.~Crammer, F.~Pereira \emph{et~al.}, ``Analysis of
  representations for domain adaptation,'' \emph{NIPS}, vol.~19, p. 137, 2007.

\bibitem{duan2012learning}
L.~Duan, D.~Xu, and I.~Tsang, ``Learning with augmented features for
  heterogeneous domain adaptation,'' \emph{arXiv}, 2012.

\bibitem{pan2008transfer}
S.~J. Pan, J.~T. Kwok, and Q.~Yang, ``Transfer learning via dimensionality
  reduction.'' in \emph{AAAI}, vol.~8, 2008, pp. 677--682.

\bibitem{li2014learning}
Y.~Li, J.~Liu, H.~Lu, and S.~Ma, ``Learning robust face representation with
  classwise block-diagonal structure,'' \emph{IEEE Trans. Information Forensics
  and Security}, vol.~9, no.~12, pp. 2051--2062, 2014.

\bibitem{yao2015semi}
T.~Yao, Y.~Pan, C.-W. Ngo, H.~Li, and T.~Mei, ``Semi-supervised domain
  adaptation with subspace learning for visual recognition,'' in \emph{ICCV},
  2015, pp. 2142--2150.

\bibitem{long2014transfer}
M.~Long, J.~Wang, G.~Ding, J.~Sun, and P.~S. Yu, ``Transfer joint matching for
  unsupervised domain adaptation,'' in \emph{CVPR}, 2014, pp. 1410--1417.

\bibitem{baktashmotlagh2014domain}
M.~Baktashmotlagh, M.~T. Harandi, B.~C. Lovell, and M.~Salzmann, ``Domain
  adaptation on the statistical manifold,'' in \emph{CVPR}, 2014, pp.
  2481--2488.

\bibitem{long2016unsupervised}
M.~Long, H.~Zhu, J.~Wang, and M.~I. Jordan, ``Unsupervised domain adaptation
  with residual transfer networks,'' in \emph{NIPS}, 2016, pp. 136--144.

\bibitem{ganin2014unsupervised}
Y.~Ganin and V.~Lempitsky, ``Unsupervised domain adaptation by
  backpropagation,'' \emph{arXiv}, 2014.

\bibitem{gretton2007kernel}
A.~Gretton, K.~M. Borgwardt, M.~Rasch, B.~Sch{\"o}lkopf, A.~J. Smola
  \emph{et~al.}, ``A kernel method for the two-sample-problem,'' \emph{NIPS},
  p. 513, 2007.

\bibitem{hu2015deep}
J.~Hu, J.~Lu, and Y.-P. Tan, ``Deep transfer metric learning,'' in \emph{CVPR},
  2015, pp. 325--333.

\bibitem{long2015learning}
M.~Long, Y.~Cao, J.~Wang, and M.~Jordan, ``Learning transferable features with
  deep adaptation networks,'' in \emph{ICML}, 2015, pp. 97--105.

\bibitem{hoffman2014continuous}
J.~Hoffman, T.~Darrell, and K.~Saenko, ``Continuous manifold based adaptation
  for evolving visual domains,'' in \emph{CVPR}, 2014, pp. 867--874.

\bibitem{Yang2017EPR}
X.~Yang, M.~Wang, R.~Hong, Q.~Tian, and Y.~Rui, ``Enhancing person
  re-identification in a self-trained subspace,'' \emph{TOMM}, vol.~13, no.~3,
  pp. 27:1--27:23, 2017.

\bibitem{kanamori2009efficient}
T.~Kanamori, S.~Hido, and M.~Sugiyama, ``Efficient direct density ratio
  estimation for non-stationarity adaptation and outlier detection,'' in
  \emph{NIPS}, 2009, pp. 809--816.

\bibitem{yang2007cross}
J.~Yang, R.~Yan, and A.~G. Hauptmann, ``Cross-domain video concept detection
  using adaptive svms,'' in \emph{ACM MM}, 2007, pp. 188--197.

\bibitem{Bruzzone2010Domain}
L.~Bruzzone and M.~Marconcini, ``Domain adaptation problems: A dasvm
  classification technique and a circular validation strategy,'' \emph{IEEE
  Trans. PAMI}, vol.~32, no.~5, pp. 770--787, 2010.

\bibitem{eda2016zhang}
L.~Zhang and D.~Zhang, ``Robust visual knowledge transfer via extreme learning
  machine-based domain adpatation,'' \emph{IEEE Trans. Image Processing},
  vol.~25, no.~3, pp. 4959--4973, 2016.

\bibitem{li2014hfa}
W.~Li, L.~Duan, D.~Xu, and I.~Tsang, ``Learning with augmented features for
  supervised and semi-supervised heterogeneous domain adaptation,'' \emph{IEEE
  Trans. PAMI}, vol.~36, no.~6, pp. 1134--1148, 2014.

\bibitem{Hoffman2014Asymmetric}
J.~Hoffman, E.~Rodner, J.~Donahue, B.~Kulis, and K.~Saenko, ``Asymmetric and
  category invariant feature transformations for domain adaptation,''
  \emph{IJCV}, vol. 109, no.~1, pp. 28--41, 2014.

\bibitem{Gong2012Geodesic}
B.~Gong, Y.~Shi, F.~Sha, and K.~Grauman, ``Geodesic flow kernel for
  unsupervised domain adaptation,'' in \emph{CVPR}, 2012, pp. 2066--2073.

\bibitem{Gopalan2011Domain}
R.~Gopalan, R.~Li, and R.~Chellappa, ``Domain adaptation for object
  recognition: An unsupervised approach,'' in \emph{Proc. ICCV}, 2011, pp.
  999--1006.

\bibitem{gretton2012jmlr}
A.~Gretton, K.~M. Borgwardt, M.~J. Rasch, B.~Scholk{\"o}pf, and A.~Smola, ``A
  kernel two-sample test,'' \emph{JMLR}, pp. 723--773, 2012.

\bibitem{iyer2014maximum}
A.~Iyer, J.~S. Nath, and S.~Sarawagi, ``Maximum mean discrepancy for class
  ratio estimation: Convergence bounds and kernel selection.'' in \emph{ICML},
  2014, pp. 530--538.

\bibitem{long2013transfer}
M.~Long, G.~Ding, J.~Wang, J.~Sun, Y.~Guo, and P.~S. Yu, ``Transfer sparse
  coding for robust image representation,'' in \emph{ICCV}, 2013, pp. 407--414.

\bibitem{jhuo2012robust}
I.-H. Jhuo, D.~Liu, D.~Lee, and S.-F. Chang, ``Robust visual domain adaptation
  with low-rank reconstruction,'' in \emph{CVPR}, 2012, pp. 2168--2175.

\bibitem{shao2014generalized}
M.~Shao, D.~Kit, and Y.~Fu, ``Generalized transfer subspace learning through
  low-rank constraint,'' \emph{IJCV}, vol. 109, no. 1-2, pp. 74--93, 2014.

\bibitem{zhang2016lsdt}
L.~Zhang, W.~Zuo, and D.~Zhang, ``Lsdt: Latent sparse domain transfer learning
  for visual adaptation,'' \emph{IEEE Trans Image Processing}, vol.~25, no.~3,
  pp. 1177--1191, 2016.

\bibitem{Zhang2016Discriminative}
L.~Zhang, S.~K. Jha, T.~Liu, and G.~Pei, ``Discriminative kernel transfer
  learning via l2,1-norm minimization,'' in \emph{IJCNN}, 2016.

\bibitem{Xu2015}
Y.~Xu, X.~Fang, J.~Wu, X.~Li, and D.~Zhang, ``Discriminative transfer subspace
  learning via low-rank and sparse representation.'' \emph{IEEE Trans Image
  Processing}, vol.~25, no.~2, pp. 850--863, 2015.

\bibitem{tzeng2015simultaneous}
E.~Tzeng, J.~Hoffman, T.~Darrell, and K.~Saenko, ``Simultaneous deep transfer
  across domains and tasks,'' in \emph{ICCV}, 2015, pp. 4068--4076.

\bibitem{glorot2011domain}
X.~Glorot, A.~Bordes, and Y.~Bengio, ``Domain adaptation for large-scale
  sentiment classification: A deep learning approach,'' in \emph{ICML}, 2011,
  pp. 513--520.

\bibitem{oquab2014learning}
M.~Oquab, L.~Bottou, I.~Laptev, and J.~Sivic, ``Learning and transferring
  mid-level image representations using convolutional neural networks,'' in
  \emph{CVPR}, 2014, pp. 1717--1724.

\bibitem{xie2015transfer}
M.~Xie, N.~Jean, M.~Burke, D.~Lobell, and S.~Ermon, ``Transfer learning from
  deep features for remote sensing and poverty mapping,'' \emph{arXiv}, 2015.

\bibitem{donahue2014decaf}
J.~Donahue, Y.~Jia, O.~Vinyals, J.~Hoffman, N.~Zhang, E.~Tzeng, and T.~Darrell,
  ``Decaf: A deep convolutional activation feature for generic visual
  recognition,'' in \emph{ICML}, 2014, pp. 647--655.

\bibitem{sharif2014cnn}
A.~Sharif~Razavian, H.~Azizpour, J.~Sullivan, and S.~Carlsson, ``Cnn features
  off-the-shelf: an astounding baseline for recognition,'' in \emph{CVPR},
  2014, pp. 806--813.

\bibitem{Tzeng2017Adversarial}
E.~Tzeng, J.~Hoffman, K.~Saenko, and T.~Darrell, ``Adversarial discriminative
  domain adaptation,'' 2017, cVPR.

\bibitem{Goodfellow2014Generative}
I.~Goodfellow, J.~Pouget-Abadie, M.~Mirza, B.~Xu, D.~Warde-Farley, S.~Ozair,
  A.~Courville, and Y.~Bengio, ``Generative adversarial nets,'' in \emph{NIPS},
  2014, pp. 2672--2680.

\bibitem{Lin2011Linearized}
Z.~Lin, R.~Liu, and Z.~Su, ``Linearized alternating direction method with
  adaptive penalty for low-rank representation,'' \emph{NIPS}, pp. 612--620,
  2011.

\bibitem{Cai2008A}
J.~F. Cai, E.~J. Candes, and Z.~Shen, ``A singular value thresholding algorithm
  for matrix completion,'' \emph{Siam Journal on Optimization}, vol.~20, no.~4,
  pp. 1956--1982, 2008.

\bibitem{Rosasco2009Iterative}
L.~Rosasco, A.~Verri, M.~Santoro, S.~Mosci, and S.~Villa, ``Iterative
  projection methods for structured sparsity regularization,''
  \emph{Computation}, 2009.

\bibitem{kanamori2009least}
T.~Kanamori, S.~Hido, and M.~Sugiyama, ``A least-squares approach to direct
  importance estimation,'' \emph{JMLR}, vol.~10, no. Jul, pp. 1391--1445, 2009.

\bibitem{wright2009robust}
J.~Wright, A.~Y. Yang, A.~Ganesh, S.~S. Sastry, and Y.~Ma, ``Robust face
  recognition via sparse representation,'' \emph{IEEE Trans. PAMI}, vol.~31,
  no.~2, pp. 210--227, 2009.

\bibitem{gaidon2014self}
A.~Gaidon, G.~Zen, and J.~A. Rodriguez-Serrano, ``Self-learning camera:
  Autonomous adaptation of object detectors to unlabeled video streams,''
  \emph{arXiv}, 2014.

\bibitem{gong2014learning}
B.~Gong, K.~Grauman, and F.~Sha, ``Learning kernels for unsupervised domain
  adaptation with applications to visual object recognition,'' \emph{IJCV},
  vol. 109, no. 1-2, pp. 3--27, 2014.

\bibitem{liu2010hierarchical}
D.~Liu, G.~Hua, and T.~Chen, ``A hierarchical visual model for video object
  summarization,'' \emph{IEEE Trans. PAMI}, vol.~32, no.~12, pp. 2178--2190,
  2010.

\bibitem{Fernando2014Unsupervised}
B.~Fernando, A.~Habrard, M.~Sebban, and T.~Tuytelaars, ``Unsupervised visual
  domain adaptation using subspace alignment,'' in \emph{ICCV}, 2014, pp.
  2960--2967.

\bibitem{saenko2010eccv}
K.~Saenko, B.~Kulis, M.~Fritz, and T.~Darrell, ``Adapting visual category
  models to new domains,'' 2011, pp. 999--1006.

\bibitem{Chang2013Robust}
S.~F. Chang, D.~T. Lee, D.~Liu, and I.~Jhuo, ``Robust visual domain adaptation
  with low-rank reconstruction,'' in \emph{CVPR}, 2013, pp. 2168--2175.

\bibitem{Krizhevsky2012ImageNet}
A.~Krizhevsky, I.~Sutskever, and G.~E. Hinton, ``Imagenet classification with
  deep convolutional neural networks,'' \emph{NIPS}, pp. 1097--1105, 2012.

\bibitem{Rate2011Columbia}
C.~Rate and C.~Retrieval, ``Columbia object image library (coil-20),''
  \emph{Computer}, 2011.

\end{thebibliography}

}
%
%

%
\begin{IEEEbiography}[{\includegraphics[width=1in,height=1.25in,clip,keepaspectratio]{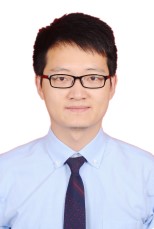}}]{Lei Zhang}
(M'14-SM'18) received his Ph.D degree in Circuits and Systems from the College of Communication Engineering, Chongqing University, Chongqing, China, in 2013. He was selected as a Hong Kong Scholar in China in 2013, and worked as a Post-Doctoral Fellow with The Hong Kong Polytechnic University, Hong Kong, from 2013 to 2015. He is currently a Professor/Distinguished Research Fellow with Chongqing University. He has authored more than 70 scientific papers in top journals and conferences, including the IEEE T-NNLS, IEEE T-IP, IEEE T-MM, IEEE T-CYB, IEEE T-IM, IEEE T-SMCA, etc. His current research interests include machine learning, pattern recognition, computer vision and intelligent systems. Dr. Zhang was a recipient of Outstanding Reviewer Award for more than 10 journals, Outstanding Doctoral Dissertation Award of Chongqing, China, in 2015, Hong Kong Scholar Award in 2014, Academy Award for Youth Innovation of Chongqing University in 2013 and the New Academic Researcher Award for Doctoral Candidates from the Ministry of Education, China, in 2012.
\end{IEEEbiography}

\begin{IEEEbiography}[{\includegraphics[width=1in,height=1.25in,clip,keepaspectratio]{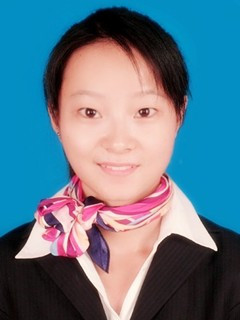}}]{Shanshan Wang}
 received BE and ME from the ChongQing University in 2010 and 2013, respectively. She is currently pursuing the Ph.D. degree at ChongQing University. Her current research interests include machine learning, pattern recognition, computer vision.
\end{IEEEbiography}


\begin{IEEEbiography}[{\includegraphics[width=1in,height=1.25in,clip,keepaspectratio]{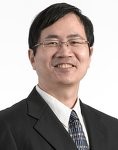}}]{Guangbin Huang}
 (M'98-SM'04) received the B.Sc. degree in applied mathematics and the M.Eng. degree in computer engineering from Northeastern University, China, in 1991 and 1994, respectively, and Ph.D. degree in electrical engineering from Nanyang Technological University, Singapore, in 1999. During undergraduate period, he also concurrently studied with the Applied Mathematics Department and Wireless Communication Department, Northeastern University, China. He serves as an Associate Editor of Neurocomputing, Neural Networks, Cognitive Computation, and the IEEE Transactions on Cybernetics. His current research interests include machine learning, computational intelligence, and extreme learning machines. He was a Research Fellow with the Singapore Institute of Manufacturing Technology, from 1998 to 2001, where he has led/implemented several key industrial projects (e.g., a Chief Designer and a Technical Leader of Singapore Changi Airport Cargo Terminal Upgrading Project). From 2001, he has been an Assistant Professor and an Associate Professor with the School of Electrical and Electronic Engineering, Nanyang Technological University, Singapore. He received the best paper award of the IEEE Transactions on Neural Networks and Learning Systems in 2013.
\end{IEEEbiography}
\begin{IEEEbiography}[{\includegraphics[width=1in,height=1.25in,clip,keepaspectratio]{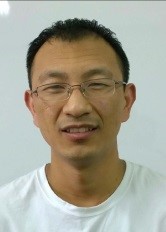}}]{Wangmeng Zuo}
received the Ph.D. degree in computer application technology from the Harbin Institute of Technology, Harbin, China, in 2007. From July 2004 to December 2004, from November 2005 to August 2006, and from July 2007 to February 2008, he was a Research Assistant at the Department of Computing, Hong Kong Polytechnic University, Hong Kong. From August 2009 to February 2010, he was a Visiting Professor in Microsoft Research Asia. He is currently an Associate Professor in the School of Computer Science and Technology, Harbin Institute of Technology. Dr. Zuo has published more than 60 papers in top tier academic journals and conferences. His current research interests include image modeling and blind restoration, discriminative learning, biometrics, and 3D vision. Dr. Zuo is an Associate Editor of the IET Biometrics. He is a senior member of the IEEE.
\end{IEEEbiography}
\begin{IEEEbiography}[{\includegraphics[width=1in,height=1.25in,clip,keepaspectratio]{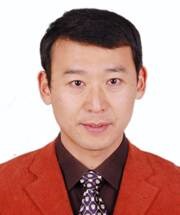}}]{Jian Yang}
received the PhD degree from Nanjing University of Science and Technology (NUST), on the subject of pattern recognition and intelligence systems in 2002. In 2003, he was a postdoctoral researcher at the University of Zaragoza. From 2004 to 2006, he was a Postdoctoral Fellow at Biometrics Centre of Hong Kong Polytechnic University. From 2006 to 2007, he was a Postdoctoral Fellow at Department of Computer Science of New Jersey Institute of Technology. Now, he is a Chang-Jiang professor in the School of Computer Science and Technology of NUST. He is the author of more than 100 scientific papers in pattern recognition and computer vision. His journal papers have been cited more than 4000 times in the ISI Web of Science, and 9000 times in the Web of Scholar Google. His research interests include pattern recognition, computer vision and machine learning. Currently, he is/was an associate editor of Pattern Recognition Letters, IEEE Trans. Neural Networks and Learning Systems, and Neurocomputing. He is a Fellow of IAPR.
\end{IEEEbiography}
\begin{IEEEbiography}[{\includegraphics[width=1in,height=1.25in,clip,keepaspectratio]{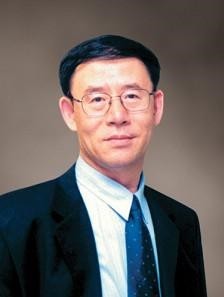}}]{David Zhang}
(F'09) graduated in Computer Science from Peking University in 1974. He received his MSc in 1982 and his PhD in 1985 in Computer Science from the Harbin Institute of Technology (HIT), respectively. From 1986 to 1988 he was a Postdoctoral Fellow at Tsinghua University and then an Associate Professor at the Academia Sinica, Beijing. In 1994 he received his second PhD in Electrical and Computer Engineering from the University of Waterloo, Ontario, Canada. He is  a Chair Professor since 2005 at the Hong Kong Polytechnic University where he is the Founding Director of the Biometrics Research Centre (UGC/CRC) supported by the Hong Kong SAR Government in 1998. He also serves as Visiting Chair Professor in Tsinghua University, and Adjunct Professor in Peking University, Shanghai Jiao Tong University, HIT, and the University of Waterloo. He is the Founder and Editor-in-Chief, International Journal of Image and Graphics (IJIG); Book Editor, Springer International Series on Biometrics (KISB); Organizer, the International Conference on Biometrics Authentication (ICBA); Associate Editor of more than ten international journals including IEEE TRANSACTIONS and so on; and the author of more than 10 books, over 300 international journal papers and 30 patents from USA/Japan/HK/China. Professor Zhang is a Croucher Senior Research Fellow, Distinguished Speaker of the IEEE Computer Society, and a Fellow of both IEEE and IAPR.
\end{IEEEbiography}



\end{document}